\documentclass[letterpaper]{article} 
\usepackage{aaai25} 
\usepackage{times}  
\usepackage{helvet}  
\usepackage{courier}  
\usepackage[hyphens]{url}  
\usepackage{graphicx} 
\usepackage{natbib}  
\usepackage{caption} 
\usepackage{algorithm}
\usepackage{algorithmic}

\usepackage{newfloat}
\usepackage{listings}
\DeclareCaptionStyle{ruled}{labelfont=normalfont,labelsep=colon,strut=off} 
\floatstyle{ruled}
\newfloat{listing}{tb}{lst}{}
\floatname{listing}{Listing}
\title{DiffExp: Efficient Exploration in Reward Fine-tuning \\for Text-to-Image Diffusion Models}
\author{
    Daewon Chae\textsuperscript{\rm 1}\footnote{Equal contribution. $^\dagger$Equal advising.}, June Suk Choi\textsuperscript{\rm 2}\footnotemark[1], Jinkyu Kim\textsuperscript{\rm 1}$^\dagger$, Kimin Lee\textsuperscript{\rm 2}$^\dagger$
}
\affiliations{
    \textsuperscript{\rm 1}Korea University, South Korea \\
    \textsuperscript{\rm 2}KAIST, South Korea \\
    cdw098@korea.ac.kr, \ w\_choi@kaist.ac.kr, \ jinkyukim@korea.ac.kr, \ kiminlee@kaist.ac.kr

}

\usepackage{bibentry}

\usepackage{xspace}
\usepackage{xcolor}

\newcommand{\metabbr}{DiffExp\xspace}

\begin{document}

\maketitle

\begin{abstract}
Fine-tuning text-to-image diffusion models to maximize rewards has proven effective for enhancing model performance. However, reward fine-tuning methods often suffer from slow convergence due to online sample generation. Therefore, obtaining diverse samples with strong reward signals is crucial for improving sample efficiency and overall performance. In this work, we introduce DiffExp, a simple yet effective exploration strategy for reward fine-tuning of text-to-image models. Our approach employs two key strategies: (a) dynamically adjusting the scale of classifier-free guidance to enhance sample diversity, and (b) randomly weighting phrases of the text prompt to exploit high-quality reward signals. We demonstrate that these strategies significantly enhance exploration during online sample generation, improving the sample efficiency of recent reward fine-tuning methods, such as DDPO and AlignProp.
\end{abstract}

%

\section{Introduction}
\begin{figure}[!t]
  \centering
  \includegraphics[width=1.0\linewidth]{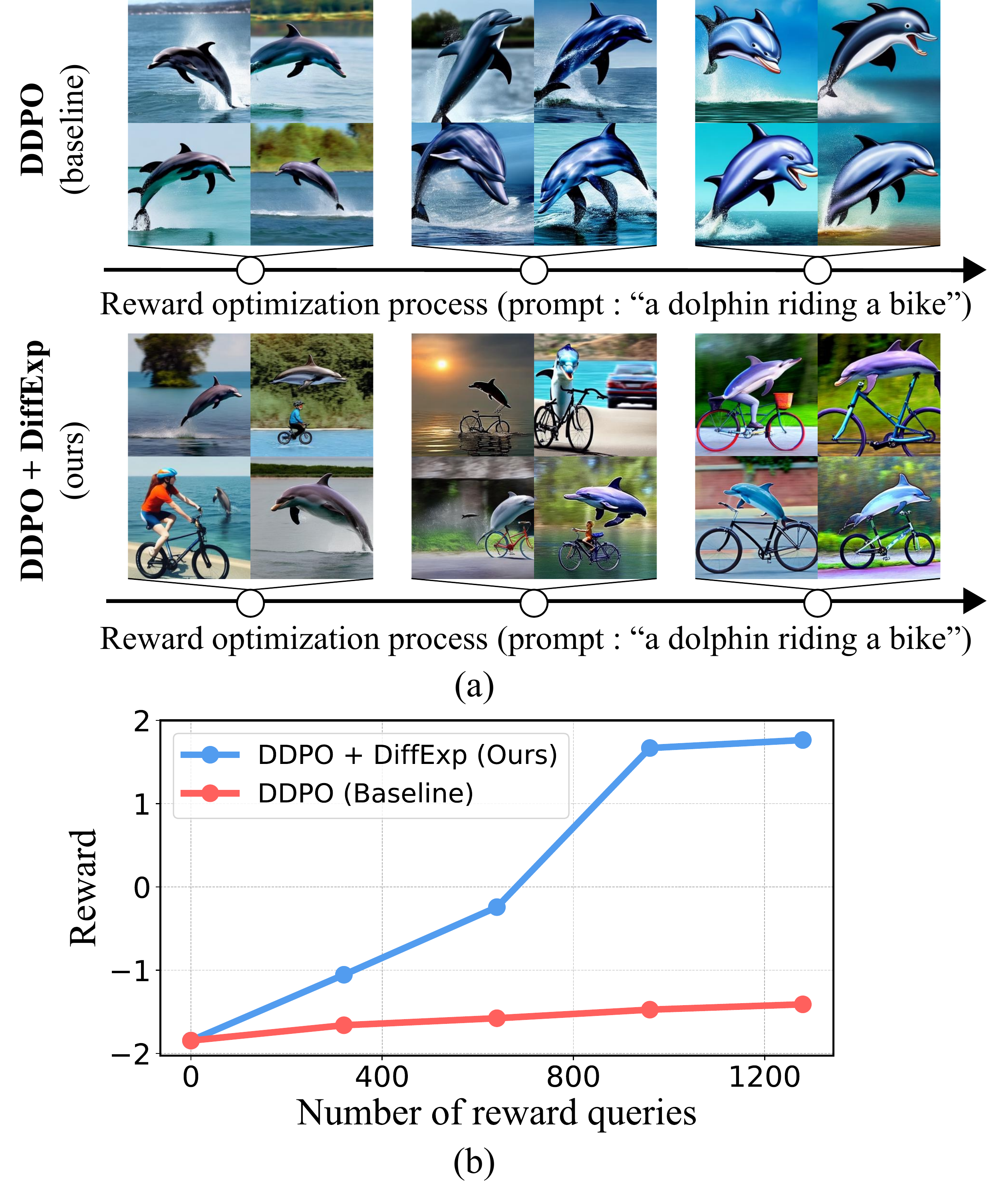}
  \caption{
  (a) Generated images during the reward optimization process with our proposed method called DiffExp given the prompt ``a dolphin riding a bike,'' which is often challenging for existing reward fine-tuning approaches, including our baseline DDPO~\cite{ddpo}. (b) We also provide corresponding reward curves against the number of reward queries, where our method indeed improves its sample efficiency during reward optimization process, capturing good reward signals for reward fine-tuning. 
  }
  \label{fig:dolphin}
\end{figure}

Reward fine-tuning~\cite{ddpo,dpok,draft,alignprop, align_t2i} has recently emerged as a powerful method for improving text-to-image diffusion models~\cite{sdxl,stable_diffusion}.
Unlike the conventional optimization strategy of likelihood maximization, this framework focuses on maximizing reward scores that measure the quality of model outputs, such as image-text alignment~\cite{pickscore,align_t2i,hps,imagereward} and image fidelity~\cite{aesthetic}.
Several methods including policy gradient~\cite{ddpo,dpok} and direct backpropagation~\cite{draft,alignprop} have been studied for reward maximization.
These methods have shown promising results in improving image-text alignment~\cite{ddpo,dpok}, reducing undesired biases~\cite{dpok}, and removing artifacts from generated images~\cite{hps}. 

Since this reward-based fine-tuning involves online sample generation, the reward optimization depends on what samples are produced during the generation process. 
In other words, if samples with good reward signals are not obtained during the generation process, the model will converge slowly due to the lack of these signals. 
To demonstrate this, we conduct reward-based fine-tuning using the single prompt ``a dolphin riding a bike'', which is known to be challenging for reward optimization~\cite{ddpo}.
In this experiment, we fine-tune Stable Diffusion 1.5~\cite{stable_diffusion} using the policy gradient method (i.e., DDPO~\cite{ddpo}) to maximize image-text alignment based on ImageReward~\cite{imagereward} scores. 
As shown in Figure~\ref{fig:dolphin}, the current method fails to find a good reward signal (i.e., images containing a bicycle) during the optimization process, resulting in trivial solutions that change the style of the image without improving image-text alignment.
This highlights the need for enhanced exploration in the sample generation process to better capture good reward signals.

In this work, we introduce \metabbr (DiffusionExplore), an exploration method designed to efficiently fine-tune text-to-image diffusion models using rewards.
Our method relies on two main strategies: dynamically controlling the scale of classifier-free guidance (CFG)~\cite{cfg} and randomly weighting certain phrases in the text prompt.
CFG is a sampling technique that balances sample diversity and fidelity.
Unlike conventional approaches that fix the CFG scale during the denoising process, we propose initially setting this scale to an extremely low value and increasing it in the later step.
This dynamic scheduling improves the diversity of the online samples while maintaining their fidelity.
To further enhance diverse sample generation, we perform additional exploration by randomly assigning weights to certain phrases within the text prompt. 
This approach creates different images where specific elements of the text prompt are emphasized randomly, thereby promoting the emergence of valuable reward signals. 
We find that these two key strategies significantly boost the effectiveness of reward-based fine-tuning (see blue curve in Figure~\ref{fig:dolphin}).

To verify the effectiveness of our exploration method, \metabbr, we integrate it with two popular reward fine-tuning techniques: policy gradient method~\cite{ddpo} and direct reward backpropagation method~\cite{alignprop}.
Specifically, we fine-tune the Stable Diffusion model~\cite{stable_diffusion} using our method to optimize various reward functions, including Aesthetic scorer~\cite{aesthetic} and PickScore~\cite{pickscore}. 
Our experiments demonstrate that \metabbr improves the sample-efficiency of both the policy gradient and reward backpropagation methods by encouraging more efficient exploration, which in turn increases the diversity of online samples.
Finally, we show that our method also significantly improves the quality of high-resolution text-to-image models like SDXL~\cite{sdxl}. We summarize our contributions as follows:

\begin{itemize}
    \item We propose \metabbr: an exploration method that improves diversity of online sample generation for efficient reward fine-tuning of text-to-image diffusion models.
    \item We show that our method enhances sample-efficiency of reward optimization in various fine-tuning methods, such as policy gradient and direct backpropagation.
    \item We demonstrate that our method is also effective when applied to the more recent SDXL model, as well as on the more challenging prompt set, DrawBench.
\end{itemize}

\begin{figure*}[tb]
  \centering
  \includegraphics[width=\linewidth]{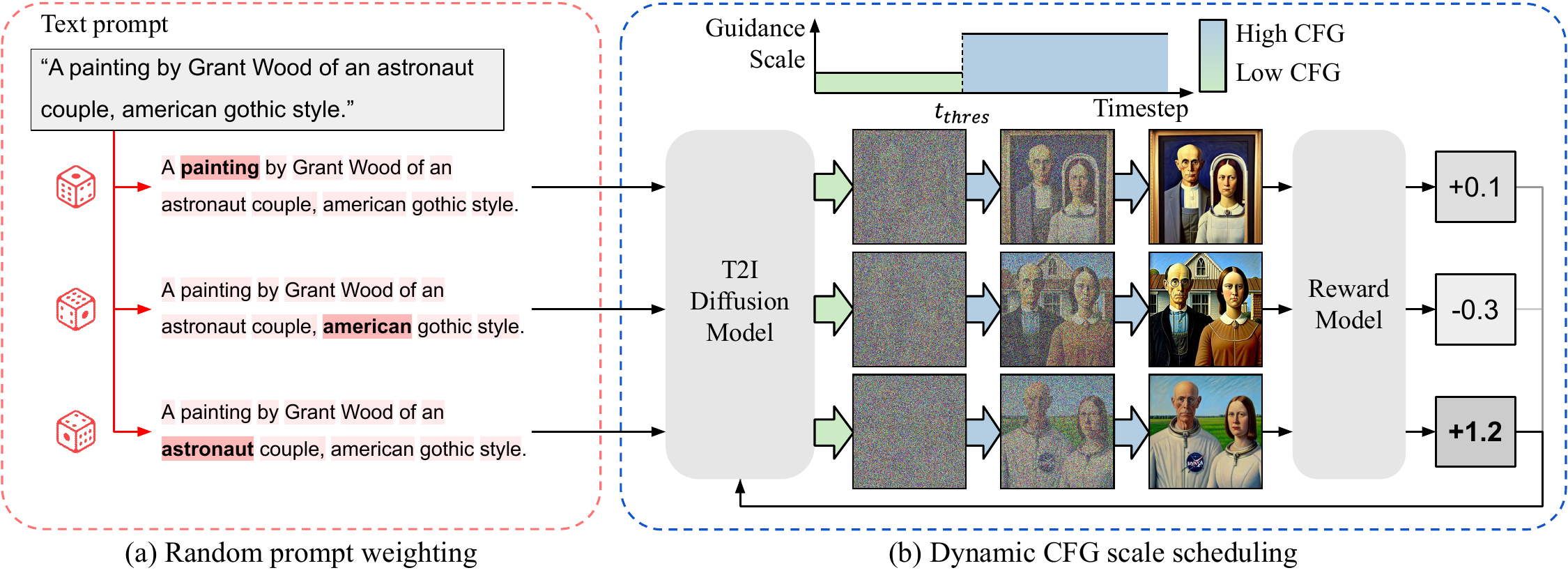}
  \caption{An overview of our proposed method called DiffExp, which consists of two main steps: (a) random prompt weighting and (b) dynamic scheduling of the CFG (classifier-free guidance) scale. In (a), word embeddings of the given prompt are randomly and differently weighted, which are then consumed by the image generation process, increasing the diversity of generated images. Further, in (b), the CFG scale of the denoising process is dynamically scheduled to control models to generate high-quality and diverse images, which is often challenging with a constantly set CFG scale. 
  }
  \label{fig:overall_method}
\end{figure*}

\section{Related Work}
\paragraph{Fine-tuning diffusion models using reward functions.}
Optimizing text-to-image diffusion models on rewards has been proven effective in supervising the final output of a diffusion model by a given reward. They are especially useful when training objectives are difficult to define given a set of images, such as human preference. Early studies~\cite{align_t2i} use supervised approaches to fine-tune the pre-trained models on images, which are weighted according to the reward function. As they are not trained online on examples, recent works~\cite{ddpo, dpok} further explore optimizing rewards using policy gradient algorithms by formulating the denoising process as a multi-step decision-making task. Though these RL approaches can flexibly train models on non-differentiable rewards, they might lose useful signals as analytic gradients of many reward functions are, in practice, available. AlignProp~\cite{alignprop} and DRaFT~\cite{draft} explored optimizing diffusion models with differentiable rewards of human preference~\cite{pickscore, hps}, effectively improving the image generation quality. 
In this work, we explore another strategy by focusing on an exploration technique toward a more efficient reward optimization process.

\paragraph{Efficient reward fine-tuning with exploration.}
In reinforcement learning (RL), where agents learn to make decisions by interacting with the environment, exploration is a crucial aspect. 
Exploration directly impacts the discovery of optimal policies, as it allows agents to gather diverse experiences. 
Intrinsic Curiosity Module (ICM)~\citep{icm} motivates exploration by rewarding actions that lead to unseen states, promoting discovery of unexplored areas of the environment.
Soft Actor-Critic (SAC)~\citep{sac} enhances exploration by employing entropy regularization, encouraging a wider range of actions. 
These works promote exploration by driving the agent to gather new experiences. 
There are works that investigate efficient reward fine-tuning using generative diffusion models as well.
A concurrent work~\citep{sergey1} proposed to endorse exploration by promoting diversity, formulating the fine-tuning of diffusion models as entropy-regularized control against pre-trained diffusion model.
Another work~\citep{sergey2} facilitates exploration by integrating an uncertainty model and KL regularization into the diffusion model tuning process.
Inspired by the success of exploration strategy in RL literature and the reward fine-tuning of diffusion models, we propose an effective exploration method for reward fine-tuning of text-to-image diffusion models.

\section{Preliminaries}
\paragraph{Text-to-image diffusion models.} 
Diffusion models~\cite{ddpm} are a class of generative models that represent the data distribution $p(x)$ by iteratively transforming Gaussian noise into data samples through a denoising process.
These models can be extended to model the conditional distribution $p(x|c)$ by incorporating an additional condition $c$.
In this work, we consider text-to-image diffusion models that utilize textual conditions to guide image generation. 
The models consist of parameterized denoising function $\epsilon_{\theta}$ and are trained by predicting added noise $\epsilon$ to the image $x$ at timestep $t$. 
Formally, given dataset $D$ comprising of image-text pairs $(x,c)$, the training objective is as follows:
\begin{equation}
\mathcal{L}_{DM}:={E}_{x, c, \epsilon, t}\left[\left\|\epsilon-\epsilon_\theta\left(x_t, t, c\right)\right\|_2^2\right],
\end{equation}
where $\epsilon$ is a Gaussian noise $\sim \mathcal{N}(0, I)$, $t$ is a timestep sampled from uniform distribution $\mathcal{U}(0, T)$, and $x_t$ is the noised image by diffusion process at a timestep $t$. At inference phase, sampling begins with drawing $x_T \sim \mathcal{N}(0, I)$, and denoising process is iteratively conducted to sample a denoised image using the estimated noise by $\epsilon_{\theta}$. 
Specifically, the estimated noise is utilized to obtain the mean of transition distribution $p(x_{t-1}|x_t, c)$ in the denoising process, as follows:
\begin{equation}
    \mu_\theta\left(x_t, t, c\right)=\frac{1}{\sqrt{\alpha_t}}\left(x_t-\frac{\beta_t}{\sqrt{1-\bar{\alpha}_t}} \epsilon_\theta\left(x_t, t, c\right)\right),
    \label{eq:ddpm_mean}
\end{equation}

\begin{equation}
p_\theta\left(x_{t-1} \mid x_{t},c\right)=\mathcal{N}\left(\mu_\theta\left(x_{t}, t, c\right), \Sigma_t\right),
\label{eq:ddpm_transition}
\end{equation}
where $\alpha_t$, $\beta_t$ are pre-defined constants used for timestep dependent denoising, $\Sigma_t$ represents covariance matrix of denoising transition, and $\bar{\alpha}_t:=\prod_{s=1}^t \alpha_s$.

\paragraph{Classifier-free guidance (CFG).} 
\citet{cfg} proposed {\em classifier-free guidance} (CFG), which serves as the basis of recent text-to-images diffusion models. In CFG, the denoising network predicts 
the noise based on the linear combination of conditional and unconditional noise estimates as follows:
\begin{equation}
\label{eq:cfg_normal}
\tilde{\epsilon}_\theta\left(x_t, t, c\right)=\epsilon_\theta\left(x_t, t\right) + w *\left(\epsilon_\theta\left(x_t, t, c\right) - \epsilon_\theta\left(x_t, t\right) \right),
\end{equation}
where $w$ is the guidance scale of CFG.
By incorporating the unconditional noise estimate into the prediction, we can achieve a trade-off between sample quality and diversity. 
Specifically, increasing the guidance weight $w$ enhances the fidelity of the samples but reduces their variety.

\section{Method}
 In this section, we introduce \metabbr (DiffusionExplore): an exploration method for efficient fine-tuning of text-to-image diffusion models using rewards. We first describe the problem setup regarding reward optimization of text-to-image diffusion models. Then, we present our approach, which promotes exploration in sample generation by scheduling CFG scale and randomly weighting a phrase of prompt. 

\subsection{Problem Formulation}
\label{sec:method_formulation}

We consider the problem of fine-tuning a text-to-image model using a reward model. The goal is to maximize the expected reward $r(x_0, c)$ for image $x_0$ generated by the model $p_\theta$ using text prompt $c$. Formally, it can be formulated as follows:
\begin{equation}
  \max_\theta E_{p(c)} E_{p_\theta\left(x_0 \mid c\right)}\left[r(x_0, c)\right],
  \label{eq:reward_obj}
\end{equation}
where $p(c)$ is the training prompt distribution and $p_\theta(\cdot|c)$ is the sample distribution for the generated image $x_0$. To achieve this goal, we consider online optimization methods that continuously generate image samples while optimizing models with rewards. Specifically, we utilize two types of reward fine-tuning methods: {\em the policy gradient method}~\cite{dpok,ddpo} and {\em direct reward backpropagation method}~\cite{draft, alignprop}. In the policy gradient method, the denoising process is defined as a multi-step Markov Decision Process, and the transition distribution $p_\theta\left(x_{t-1} \mid x_{t},c\right)$ (described in Equation~\ref{eq:ddpm_transition}) is treated as a policy. Based on this framework, the gradient of the objective in Equation 1 is computed as follows to update the model:
\begin{equation}
E_{p(c),p_\theta(x_0|c)}\left[\sum_{t=1}^T \nabla_\theta \log p_\theta\left(x_{t-1}| x_t, c\right)r\left(x_0, c\right)\right].
    \label{eq:diffusion_policy}
\end{equation}
As an alternative approach, the direct reward backpropagation method assumes that the reward model is differentiable and directly backpropagates the gradient of the reward function through the denoising process to optimize the objective in Equation ~\ref{eq:reward_obj}.

\subsection{Dynamic Scheduling of CFG Scale}

\begin{figure}[t]
  \centering
  \includegraphics[width=1.0\linewidth]{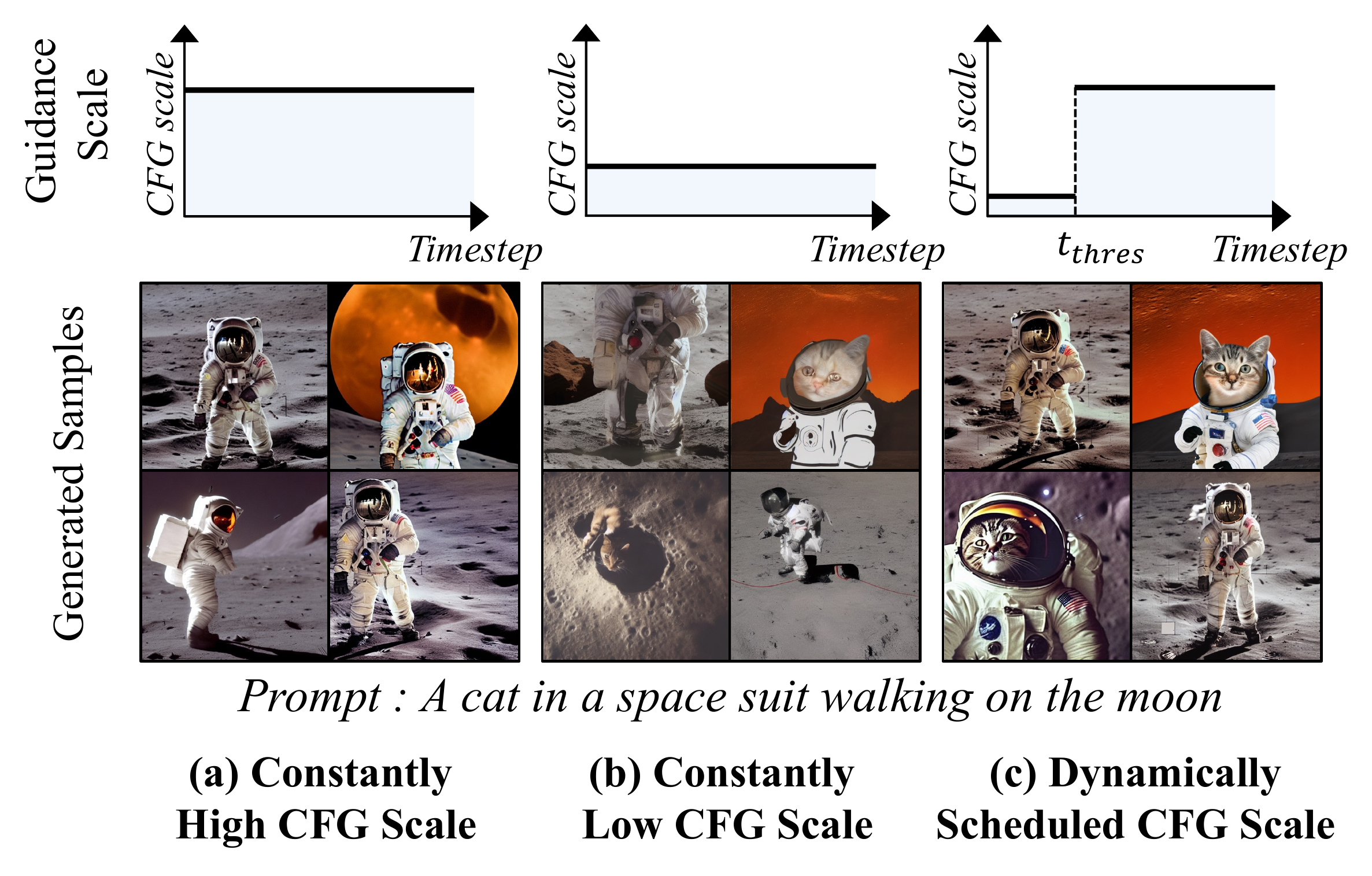}
  \caption{Comparison of the generated images with different CFG scale scheduling strategies: (a) constantly high CFG scale, (b) constantly low CFG scale, and (c) dynamically scheduled CFG scale. We observe that a model often shows increased sample diversity with a low CFG scale but suffers from degraded image quality, which is generally the opposite with a high CFG scale. Instead, dynamically scheduling the CFG scale balances sample diversity and image quality. 
  }
  \label{fig:cfg_scale_comparison}
\end{figure}

To effectively discover good reward signals, it is crucial to generate diverse image samples through exploration. To achieve this, we propose dynamically scheduling the CFG scale of the denoising process during online optimization. 
As detailed in the preliminaries, the CFG scale controls the trade-off between sample quality and diversity: a high CFG scale yields high-quality but low-diversity samples, whereas a low CFG scale promotes diversity at the cost of reduced quality (see Figure \ref{fig:cfg_scale_comparison}).
To optimize this balance, we start with a low CFG scale during the early stages of the denoising process and switch to a high CFG scale after the $t_{thres}$ denoising step, as follows:
$$\tilde{\epsilon}_\theta\left(x_t, t, c\right) = \epsilon_\theta\left(x_t, t\right) + w(t)\cdot\left(\epsilon_\theta\left(x_t, t, c\right) - \epsilon_\theta\left(x_t, t\right)\right),$$
where
\begin{equation}
    w(t) = \left\{ 
    \begin{array}{lll}
        w_l & \textit{if} &  t > t_{thres}, \\
        w_h & \textit{if} & t \leq t_{thres},
    \end{array}
    \right.
\end{equation}
and $w_l$, $w_h$ ($w_l < w_h$) are the CFG scale values. Note that in the denoising process, we get Gaussian noise at $t=T$ while the final image is at $t=0$. We set $w_l$ to an extremely low value and $w_h$ to an ordinary CFG value (i.e., 5.0 or 7.5).
This dynamic scheduling adaptively balances between image quality and diversity, allowing for generating diverse image samples without sacrificing overall sample quality.

\subsection{Random Prompt Weighting}

To further promote diverse sample generation, we propose an additional exploration method that alters text prompts.
Specifically, we increase the weight of a random word (i.e. token) in the text prompt embedding. 
This adjustment emphasizes the selected word in the generated image, leading to variations where different elements are highlighted.
Consider the text prompt \texttt{"A dolphin riding a bike"} for example, which typically results in images featuring only a dolphin, with the bike often omitted.
By increasing the weight of the word \texttt{"bike"}, there is a high chance that text-to-image diffusion models generate the image containing a bike.
This leads to more complete representations of the original prompt, thereby achieving higher prompt alignment rewards and enhancing image diversity.

Formally, given a text embedding $c$ with $N_{words}$ words, we choose a random word with index $0 \leq i < N_{words}$. Then, we increase the weight of the chosen word as follows:
\begin{equation}
c[i] \longleftarrow c_{null}+w_{prompt} *\left(c[i]-c_{null}\right),
\end{equation}
where $c_{null}$ is the text embedding that corresponds to the empty text (``''), and $w_{prompt}$ is the prompt weight. 
We find that sampling $w_{prompt}$ randomly from $\mathcal{U}(1, 1.2)$ every time is generally successful.

\begin{figure*}[t]
  \centering
  \includegraphics[width=1.0\linewidth]{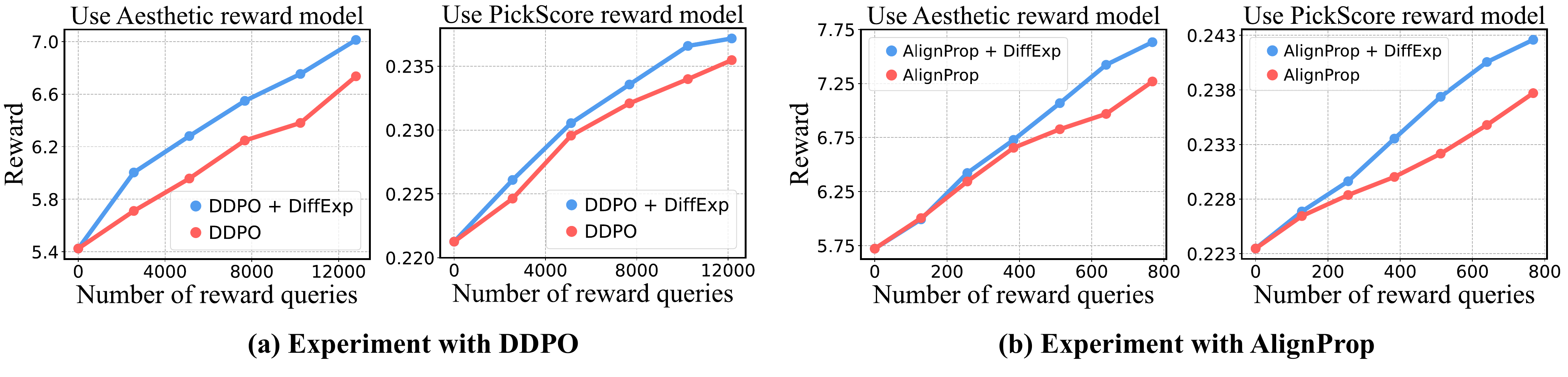}
  \caption{Reward curves for training prompts. At each checkpoint, we generate 10 images per seen prompt and use the average of their reward scores. Our sampling method is employed only during fine-tuning, not for plotting this curve.}
  \label{fig:seen_curve}
\end{figure*}

\section{Experiments}
We design our experiments to investigate the following questions:
\begin{enumerate}
    \item Can our exploration method enhance the sample-efficiency of reward fine-tuning methods?
    \item Can our exploration method improve the quality of generated samples?
    \item Can our exploration method exhibit generalization ability for unseen prompts that are not used during fine-tuning? 
\end{enumerate}

\begin{figure*}[!t]
  \centering
  \includegraphics[width=\linewidth]{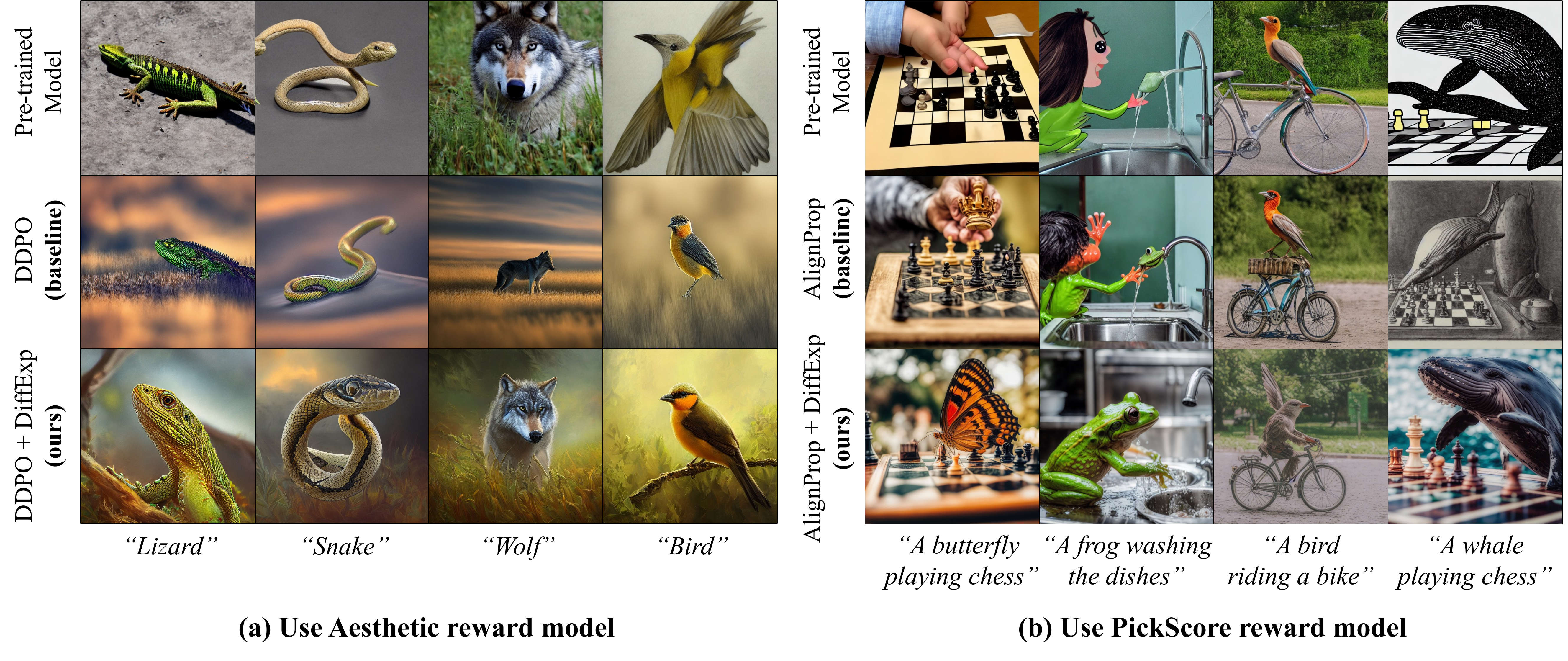}
  \caption{Generated samples from baselines and ours with (a) Aesthetic and (b) PickScore reward models. Notably, ours generates images with comparably high aesthetic quality (see (a)) and produces images with better image-text alignment given the prompts (see (b)). Note that images in the same column are generated with the same random seed. 
  }
  \label{fig:sample_quality}
\end{figure*}

\paragraph{Baselines.}
As discussed in the Problem Formulation section, we evaluate the effectiveness of our exploration method using two types of state-of-the-art reward fine-tuning methods: policy gradient method~\cite{dpok,ddpo} and direct reward backpropagation method~\cite{draft,alignprop}. We combine our approach with each of these methods and compare the performance with the original methods. Specifically, we use (1) DDPO~\cite{ddpo} for the policy gradient method and (2) AlignProp~\cite{alignprop} for direct reward backpropagation method.

\paragraph{Reward functions.}
To evaluate our method across different types of rewards, we conduct experiments using two distinct reward functions. First, we utilize an Aesthetic Score~\cite{aesthetic}, which is trained to predict the aesthetic quality of images. Following the baseline~\cite{ddpo, alignprop}, we use 45 animal names as training prompts for the aesthetic quality task. Second, in order to improve image-text alignment, we employ PickScore~\cite{pickscore}, an open-source reward model trained on a large-scale human feedback dataset. Based on the baseline~\cite{ddpo}, we use a total of 135 prompts for the image-text alignment task, combining 45 different animal names with 3 different activities (e.g., ``a monkey washing the dishes''). We provide the entire set of prompts used for training in the supplementary materials.

\paragraph{Implementation details.}
In our experiment, we use Stable Diffusion v1.5~\cite{stable_diffusion} as the pre-trained text-to-image diffusion model. Following the baselines, we employ Low-Rank Adaptation (LoRA)~\cite{lora} rather than fine-tuning the weights of pre-trained text-to-image model. As for scheduling exploration, we apply our exploration method only up to the three-fourths of the entire fine-tuning. We provide more implementation details in the supplemental material.

\begin{figure*}[!t]
  \centering
  \includegraphics[width=1\linewidth]{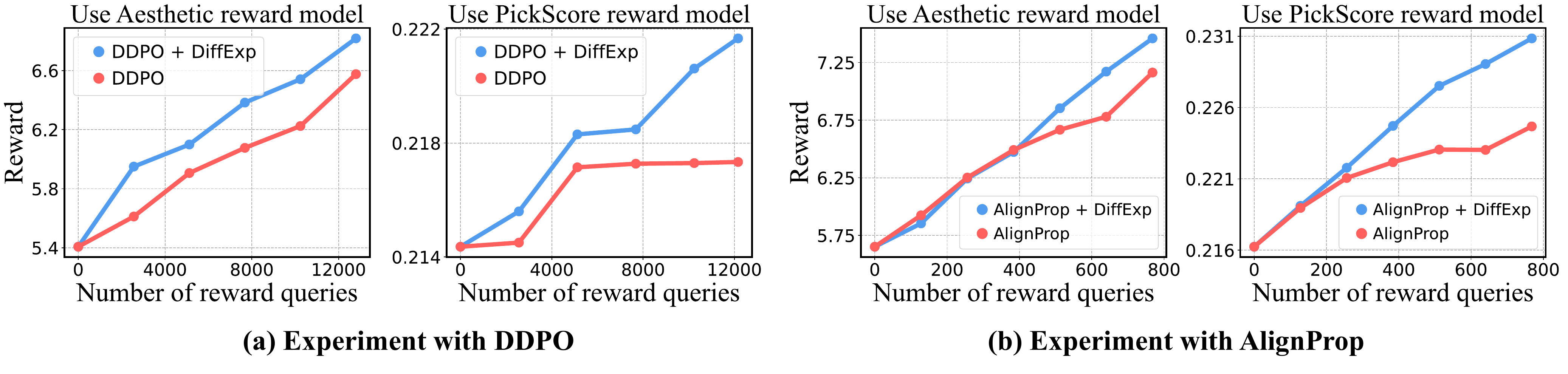}
  \caption{Reward curve for unseen prompts. At each checkpoint, we generate 10 images per unseen prompt and use the average of their reward scores. Our sampling method is employed only during fine-tuning, not for plotting this curve.}
  \label{fig:unseen_curve}
\end{figure*}

\subsection{Experimental Results}
\paragraph{Sample efficiency in reward fine-tuning.}
We begin by conducting experiments and comparing our method with DDPO and AlignProp across two reward functions: Aesthetic and PickScore. Figure~\ref{fig:seen_curve} presents the results, highlighting the achieved reward scores and sample efficiency. Notably, our method outperforms the baselines in terms of reward scores, while demonstrating better sample efficiency. Specifically, our approach required approximately 20\% fewer samples to achieve the same reward score as DDPO and AlignProp.

\begin{figure}[t]
  \centering
  \includegraphics[width=1.0\linewidth]{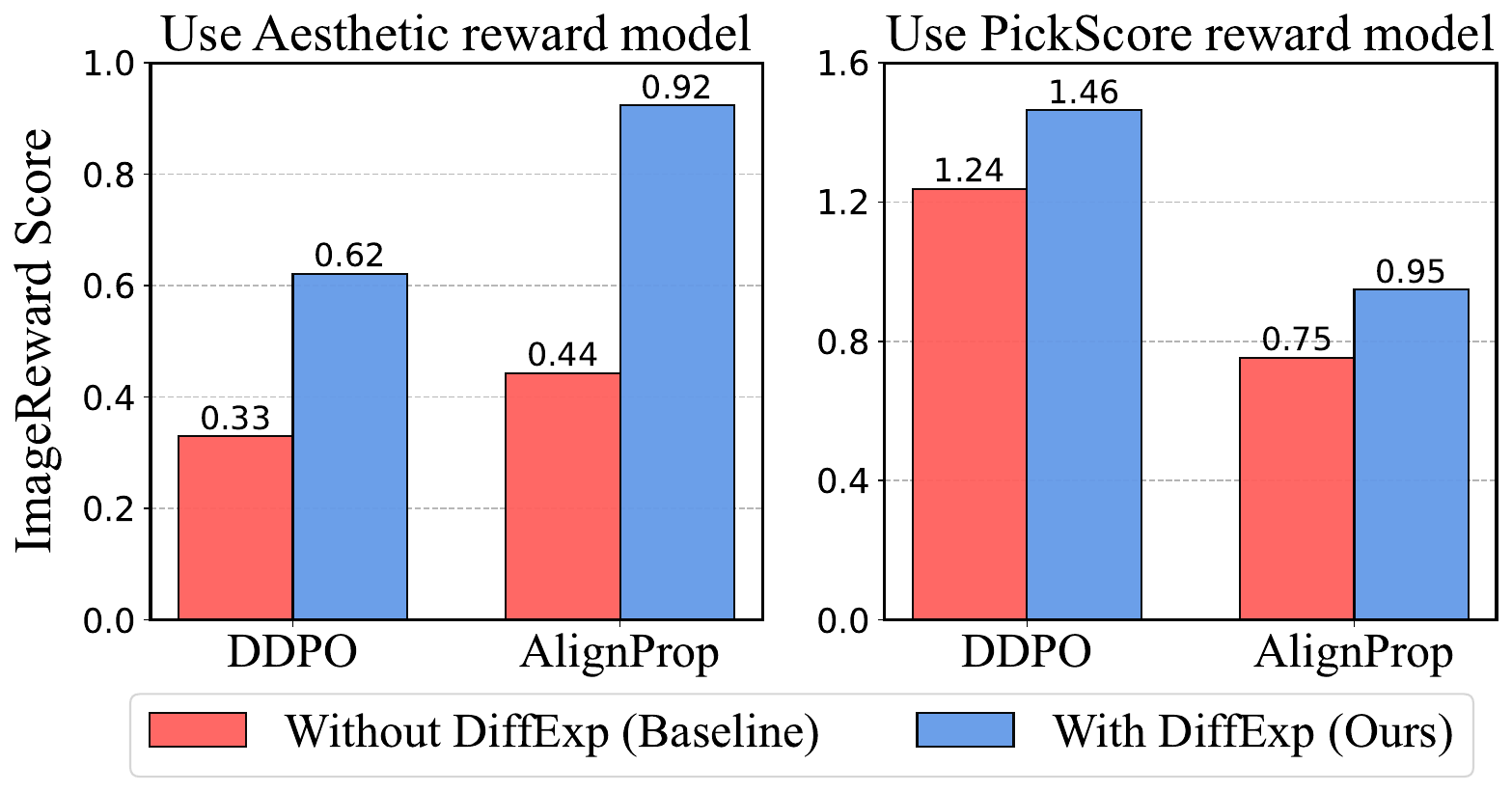}
  \caption{
  Comparison of an unseen ImageReward~\cite{imagereward} score after reward fine-tuning with and without our proposed \metabbr on our baselines, such as DDPO and AlignProp. The proposed method \metabbr achieves higher reward compared to all other baselines. 
  Note that models are optimized with either Aesthetic or PickScore as a reward function. 
  }
  \label{fig:imgreward_score}
\end{figure}

\paragraph{Qualitative comparison of image quality.}
We provide visual examples generated by each method in Figure~\ref{fig:sample_quality}. 
With the Aesthetic reward function, we observe that our method generates images with relatively high aesthetic quality, which aligns with the intent of reward function (see Figure ~\ref{fig:sample_quality} (a)). Notably, with the PickScore reward function, we observe that our method plays an important role in generating images with high image-text alignment (see Figure ~\ref{fig:sample_quality} (b)). For example, given a text prompt ``A butterfly playing chess,'' all baselines result in generating a human (not a butterfly) playing chess (see 1st column). We provide more diverse generated samples in the supplemental materials.

\paragraph{Quantitative comparison of image quality.}
To further quantitatively evaluate the sample quality, we use an ``unseen'' reward function called ImageReward~\cite{imagereward} (which is trained on a large-scale human preference dataset) that was not used during reward fine-tuning (i.e., not in the optimization objectives). In Figure~\ref{fig:imgreward_score}, we observe that ours clearly achieves higher ImageReward scores compared to baselines, such as DDPO and AlignProp fine-tuned with Aesthetic and PickScore reward functions. This demonstrates that our \metabbr indeed improves the overall image quality rather than merely increasing the reward through reward hacking~\cite{scaling_law_overopt, kim2024confidenceaware}.

\begin{figure}[t]
  \centering
  \includegraphics[width=1.0\linewidth]{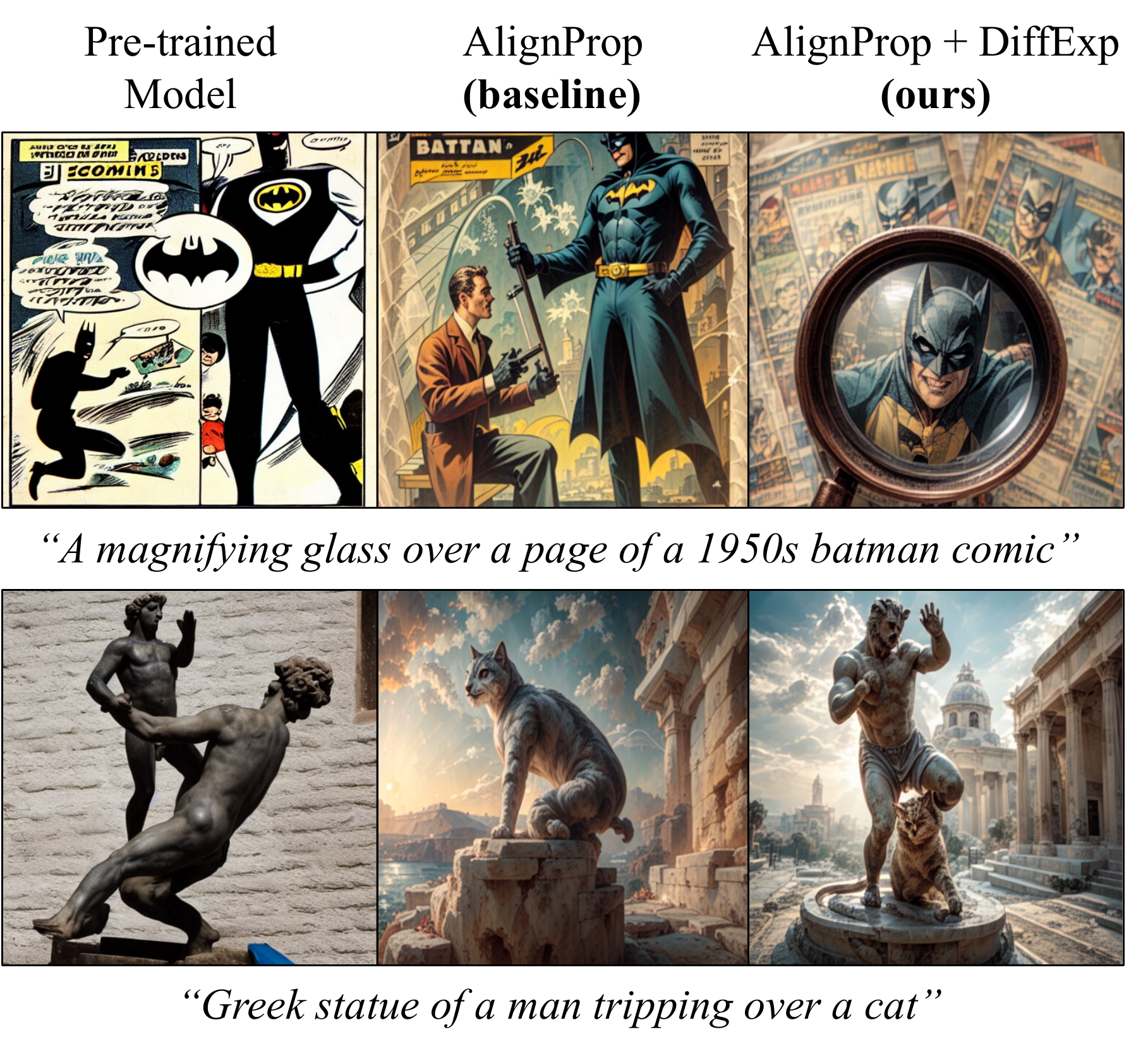}
  \caption{Generated image samples with DrawBench prompts, which are known to be challenging for the current text-to-image models. Our method successfully generates images with high sample quality and image-text alignment, capturing contexts of the given prompts such as ``a magnifying glass'' and ``a man tripping over a cat''. 
  }
  \label{fig:drawbench_img}
\end{figure}

\paragraph{Generalization to unseen prompts.}
We also investigate the model's capacity for generalization to new prompts, which were not used during reward fine-tuning. Following the standard evaluation protocol~\cite{ddpo, alignprop}, we consider a novel test set of animal names (for models fine-tuned with Aesthetic reward model) and activities (for models fine-tuned with PickScore reward model) that were not encountered during the training phase. A more detailed explanation of the experimental setup is provided in the supplemental material. As shown in Figure~\ref{fig:unseen_curve}, which shows reward curves to compare ours with baselines, a model with \metabbr consistently outperforms the baseline models in terms of rewards and sample efficiency, confirming its ability to generalize well to unseen prompts.  

\begin{figure*}[!t]
  \centering
  \includegraphics[width=1.0\linewidth]{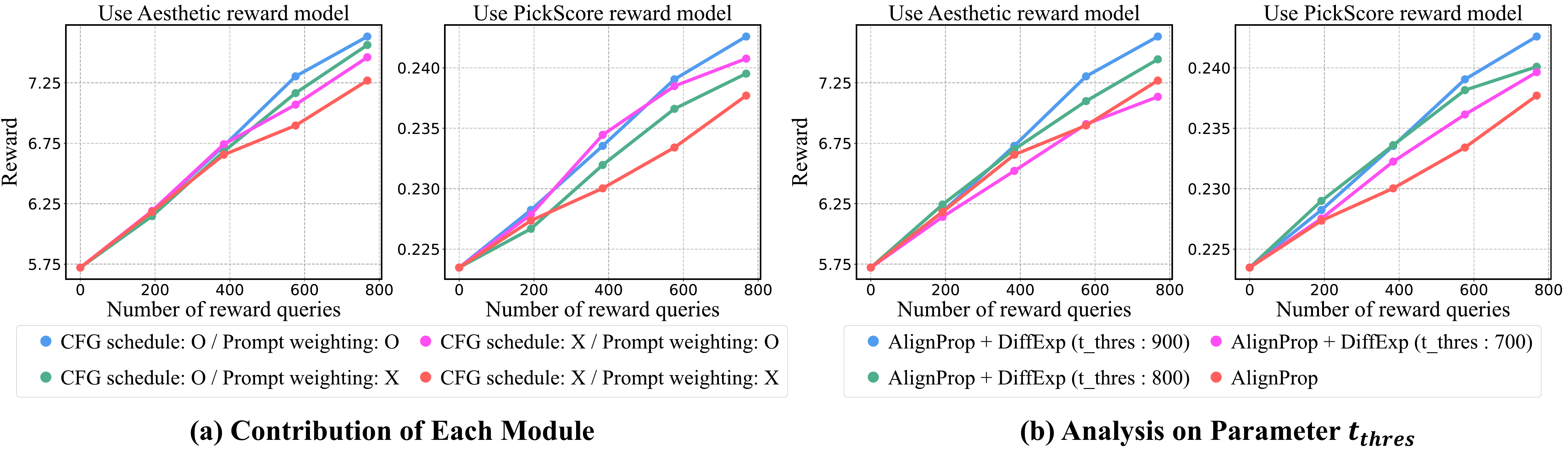}
  \caption{
  The results of our ablation study represent (a) the effect of each main module and (b) the impact of different hyper-parameter values of $t_{thres}$. 
  }
  \label{fig:analysis_curve}
\end{figure*}

\begin{figure}[t]
  \centering
  \includegraphics[width=1.0\linewidth]{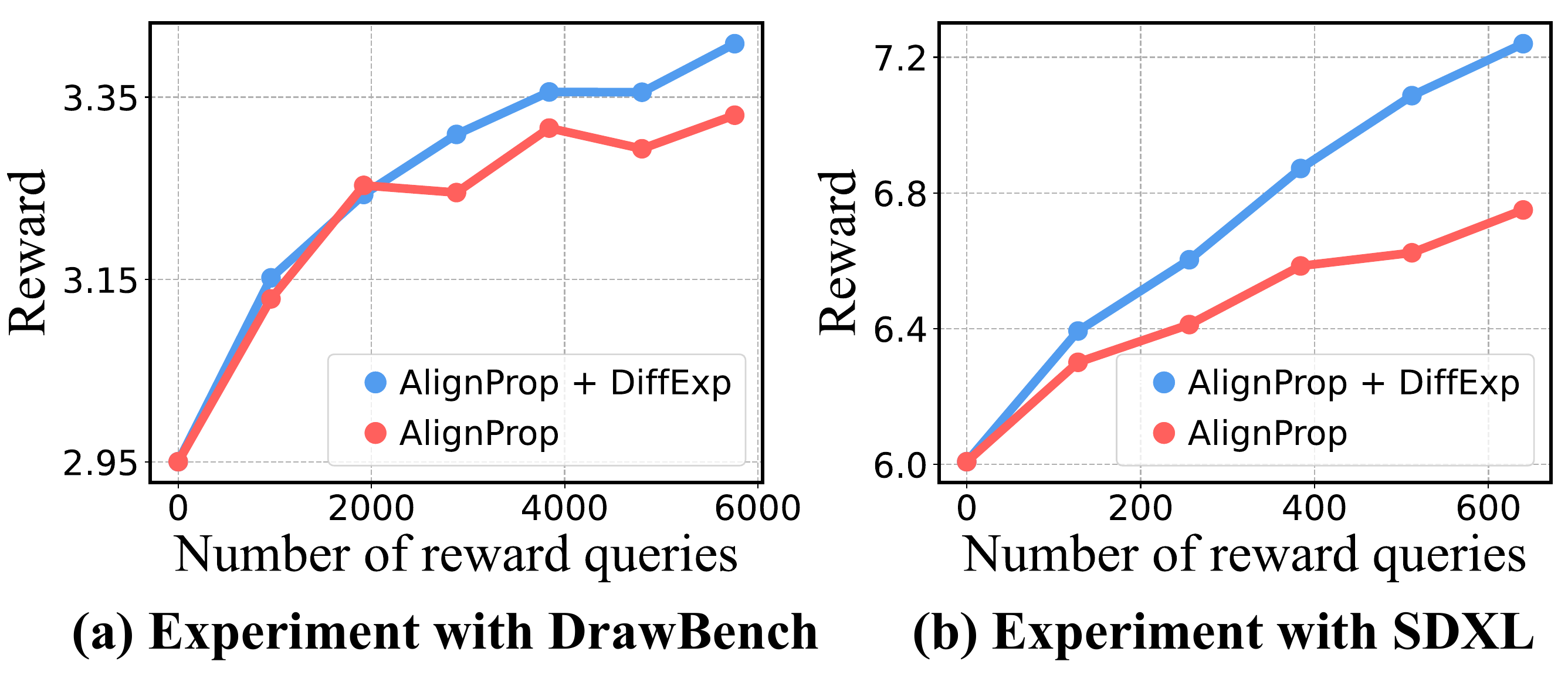}
  \caption{Reward curves for our experiments with (a) DrawBench and (b) Stable Diffusion XL (SDXL). The former evaluates the ability to generate images with challenging prompts, while the latter evaluates the adaptability to the current SOTA model.}
  \label{fig:application_curve}
\end{figure}

\paragraph{Experiment with challenging prompts.} 
Our proposed DiffExp provides an effective way to encourage the generation of diverse samples during reward fine-tuning. Thus, we argue that DiffExp is advantageous for learning more complicated prompts that are often challenging with conventional approaches, such as AlignProp. To evaluate this, we collect 58 challenging prompts from DrawBench~\cite{imagen}, which is well-known to be challenging for the current text-to-image models. We use AlignProp as our baseline model and fine-tune with an ensemble of HPS~\cite{hps}, Aesthetic, and PickScore reward models, which is reported as effective in previous studies~\cite{draft,prdp}, especially for complicated prompts. As shown in Figure~\ref{fig:drawbench_img}, the pre-trained model and AlignProp struggle to generate images aligned well with the given prompts. For example, such models are often overfitted to certain phrases (e.g., ``Batman''), ignoring other contexts (e.g., ``magnifying glass''). In contrast, our proposed DiffExp successfully generates samples with high image quality and high text fidelity (see the last column), confirming that our exploration method effectively improves reward fine-tuning with complicated (and challenging) prompts. We provide more diverse examples in supplemental material. We further provide reward curves for this experiment in Figure~\ref{fig:application_curve} (a), which is consistent with our previous analysis, demonstrating the effectiveness of our approach even to the challenging scenarios.

\paragraph{Experiment with Stable Diffusion XL (SDXL).}
In order to verify the usefulness of \metabbr with state-of-the-art models, we fine-tune SDXL~\citep{sdxl} to maximize Aesthetic Score on animal prompts using \metabbr. 
As shown in Figure~\ref{fig:application_curve}, using AlignProp with \metabbr achieves a significantly higher maximum reward value as well as better sample efficiency compared to AlignProp without \metabbr. This trend is in accordance with our experiments using Stable Diffusion v1.5, which are presented in Figure \ref{fig:seen_curve}. We include the visual examples in the supplementary material.

\paragraph{Ablation studies.}
We conduct an ablation study to see the effect of each proposed module: (i) random prompt weighting and (ii) dynamic scheduling of the CFG scale. In Figure~\ref{fig:analysis_curve} (a), we provide reward curves with variants of our models with and without those modules. We observe that all our variants generally perform better than our baseline, AlignProp, in terms of rewards and sample efficiency, while using those modules together provides the best results. This trend is consistent with different reward models, such as Aesthetic and PickScore. Further, we experiment with different values of hyper-parameter, $t_{thres}$, which determines how long the CFG scale should be maintained to a low value. In Figure~\ref{fig:analysis_curve} (b), we provide reward curves for variants of our models with $t_{thres}=\{900, 800, 700\}$. Note that in the denoising process, we have Gaussian noise at $t=1000$ while the final image is at $t=0$. We observe that higher $t_{thres}$ generally provides better results in terms of rewards, with the best performance with $t_{thres}=900$, and lowering it degrades the image quality (consistent with our analysis in Figure~\ref{fig:cfg_scale_comparison}). 

\section{Conclusion}
In this work, we propose \metabbr, an exploration method for promoting diverse reward signals during the reward optimization of text-to-image models. We demonstrate that adjusting the CFG scale of denoising process and randomly weighting certain phrase within prompts can serve as effective exploration strategy by improving the diversity of online sample generation. 
We conduct extensive experiments to verify that \metabbr improves both sample efficiency and generated image quality, and demonstrate this across various reward fine-tuning methods such as DDPO or AlignProp.
Furthermore, we conduct analysis using more advanced prompt sets such as DrawBench, and apply our method to the more advanced diffusion models such as SDXL, both of which result in significant performance improvements.

\clearpage
\newpage
\section{Acknowledgements}
This work was supported by Artificial intelligence industrial convergence cluster development project funded by the Ministry of Science and ICT(MSIT, Korea)\&Gwangju Metropolitan City (10\%) and supported by Institute of Information \& communications Technology Planning \& Evaluation(IITP) under the Leading Generative AI Human Resources Development(IITP-2025-RS-2024-00397085, 20\%), Adaptive Personality for Intelligent Agents (RS-2022-II220043, 20\%), ICT Creative Consilience Program (IITP-2025-2020-0-01819, 10\%), Global AI Frontier Lab (RS-2024-00509279, 20\%), Artificial Intelligence Graduate School Program(KAIST) (RS-2019-II190075 , 10\%), and Development and Study of AI Technologies to Inexpensively Conform to Evolving Policy on Ethics (RS-2022-II220184, 2022-0-00184, 10\%) grant funded by the Korea government(MSIT). This work was also supported by the Hyundai Motor Chung Mong-Koo Foundation.

\bibliography{aaai25}

\begin{thebibliography}{23}
\providecommand{\natexlab}[1]{#1}

\bibitem[{Black et~al.(2024)Black, Janner, Du, Kostrikov, and Levine}]{ddpo}
Black, K.; Janner, M.; Du, Y.; Kostrikov, I.; and Levine, S. 2024.
\newblock Training Diffusion Models with Reinforcement Learning.
\newblock In \emph{The Twelfth International Conference on Learning Representations}.

\bibitem[{Clark et~al.(2024)Clark, Vicol, Swersky, and Fleet}]{draft}
Clark, K.; Vicol, P.; Swersky, K.; and Fleet, D.~J. 2024.
\newblock Directly Fine-Tuning Diffusion Models on Differentiable Rewards.
\newblock In \emph{The Twelfth International Conference on Learning Representations}.

\bibitem[{Deng et~al.(2024)Deng, Wang, Wei, Grundmann, and Hou}]{prdp}
Deng, F.; Wang, Q.; Wei, W.; Grundmann, M.; and Hou, T. 2024.
\newblock PRDP: Proximal Reward Difference Prediction for Large-Scale Reward Finetuning of Diffusion Models.
\newblock \emph{arXiv preprint arXiv:2402.08714}.

\bibitem[{Fan et~al.(2024)Fan, Watkins, Du, Liu, Ryu, Boutilier, Abbeel, Ghavamzadeh, Lee, and Lee}]{dpok}
Fan, Y.; Watkins, O.; Du, Y.; Liu, H.; Ryu, M.; Boutilier, C.; Abbeel, P.; Ghavamzadeh, M.; Lee, K.; and Lee, K. 2024.
\newblock Reinforcement learning for fine-tuning text-to-image diffusion models.
\newblock \emph{Advances in Neural Information Processing Systems}, 36.

\bibitem[{Gao, Schulman, and Hilton(2023)}]{scaling_law_overopt}
Gao, L.; Schulman, J.; and Hilton, J. 2023.
\newblock Scaling laws for reward model overoptimization.
\newblock In \emph{International Conference on Machine Learning}, 10835--10866. PMLR.

\bibitem[{Haarnoja et~al.(2018)Haarnoja, Zhou, Abbeel, and Levine}]{sac}
Haarnoja, T.; Zhou, A.; Abbeel, P.; and Levine, S. 2018.
\newblock Soft Actor-Critic: Off-Policy Maximum Entropy Deep Reinforcement Learning with a Stochastic Actor.

\bibitem[{Ho, Jain, and Abbeel(2020)}]{ddpm}
Ho, J.; Jain, A.; and Abbeel, P. 2020.
\newblock Denoising diffusion probabilistic models.
\newblock \emph{Advances in neural information processing systems}, 33: 6840--6851.

\bibitem[{Ho and Salimans(2022)}]{cfg}
Ho, J.; and Salimans, T. 2022.
\newblock Classifier-free diffusion guidance.
\newblock \emph{arXiv preprint arXiv:2207.12598}.

\bibitem[{Hu et~al.(2022)Hu, yelong shen, Wallis, Allen-Zhu, Li, Wang, Wang, and Chen}]{lora}
Hu, E.~J.; yelong shen; Wallis, P.; Allen-Zhu, Z.; Li, Y.; Wang, S.; Wang, L.; and Chen, W. 2022.
\newblock Lo{RA}: Low-Rank Adaptation of Large Language Models.
\newblock In \emph{International Conference on Learning Representations}.

\bibitem[{Kim et~al.(2024)Kim, Jeong, An, Ghavamzadeh, Dvijotham, Shin, and Lee}]{kim2024confidenceaware}
Kim, K.; Jeong, J.; An, M.; Ghavamzadeh, M.; Dvijotham, K.~D.; Shin, J.; and Lee, K. 2024.
\newblock Confidence-aware Reward Optimization for Fine-tuning Text-to-Image Models.
\newblock In \emph{The Twelfth International Conference on Learning Representations}.

\bibitem[{Kirstain et~al.(2024)Kirstain, Polyak, Singer, Matiana, Penna, and Levy}]{pickscore}
Kirstain, Y.; Polyak, A.; Singer, U.; Matiana, S.; Penna, J.; and Levy, O. 2024.
\newblock Pick-a-pic: An open dataset of user preferences for text-to-image generation.
\newblock \emph{Advances in Neural Information Processing Systems}, 36.

\bibitem[{Lee et~al.(2023)Lee, Liu, Ryu, Watkins, Du, Boutilier, Abbeel, Ghavamzadeh, and Gu}]{align_t2i}
Lee, K.; Liu, H.; Ryu, M.; Watkins, O.; Du, Y.; Boutilier, C.; Abbeel, P.; Ghavamzadeh, M.; and Gu, S.~S. 2023.
\newblock Aligning text-to-image models using human feedback.
\newblock \emph{arXiv preprint arXiv:2302.12192}.

\bibitem[{Pathak et~al.(2017)Pathak, Agrawal, Efros, and Darrell}]{icm}
Pathak, D.; Agrawal, P.; Efros, A.~A.; and Darrell, T. 2017.
\newblock Curiosity-driven Exploration by Self-supervised Prediction.

\bibitem[{Podell et~al.(2023)Podell, English, Lacey, Blattmann, Dockhorn, M{\"u}ller, Penna, and Rombach}]{sdxl}
Podell, D.; English, Z.; Lacey, K.; Blattmann, A.; Dockhorn, T.; M{\"u}ller, J.; Penna, J.; and Rombach, R. 2023.
\newblock Sdxl: Improving latent diffusion models for high-resolution image synthesis.
\newblock \emph{arXiv preprint arXiv:2307.01952}.

\bibitem[{Prabhudesai et~al.(2023)Prabhudesai, Goyal, Pathak, and Fragkiadaki}]{alignprop}
Prabhudesai, M.; Goyal, A.; Pathak, D.; and Fragkiadaki, K. 2023.
\newblock Aligning text-to-image diffusion models with reward backpropagation.
\newblock \emph{arXiv preprint arXiv:2310.03739}.

\bibitem[{Rombach et~al.(2022)Rombach, Blattmann, Lorenz, Esser, and Ommer}]{stable_diffusion}
Rombach, R.; Blattmann, A.; Lorenz, D.; Esser, P.; and Ommer, B. 2022.
\newblock High-resolution image synthesis with latent diffusion models.
\newblock In \emph{Proceedings of the IEEE/CVF conference on computer vision and pattern recognition}, 10684--10695.

\bibitem[{Saharia et~al.(2022)Saharia, Chan, Saxena, Li, Whang, Denton, Ghasemipour, Gontijo~Lopes, Karagol~Ayan, Salimans et~al.}]{imagen}
Saharia, C.; Chan, W.; Saxena, S.; Li, L.; Whang, J.; Denton, E.~L.; Ghasemipour, K.; Gontijo~Lopes, R.; Karagol~Ayan, B.; Salimans, T.; et~al. 2022.
\newblock Photorealistic text-to-image diffusion models with deep language understanding.
\newblock \emph{Advances in Neural Information Processing Systems}, 35: 36479--36494.

\bibitem[{Schuhmann et~al.(2022)Schuhmann, Beaumont, Vencu, Gordon, Wightman, Cherti, Coombes, Katta, Mullis, Wortsman et~al.}]{aesthetic}
Schuhmann, C.; Beaumont, R.; Vencu, R.; Gordon, C.; Wightman, R.; Cherti, M.; Coombes, T.; Katta, A.; Mullis, C.; Wortsman, M.; et~al. 2022.
\newblock Laion-5b: An open large-scale dataset for training next generation image-text models.
\newblock \emph{Advances in Neural Information Processing Systems}, 35: 25278--25294.

\bibitem[{Uehara et~al.(2024{\natexlab{a}})Uehara, Zhao, Black, Hajiramezanali, Scalia, Diamant, Tseng, Biancalani, and Levine}]{sergey1}
Uehara, M.; Zhao, Y.; Black, K.; Hajiramezanali, E.; Scalia, G.; Diamant, N.~L.; Tseng, A.~M.; Biancalani, T.; and Levine, S. 2024{\natexlab{a}}.
\newblock Fine-Tuning of Continuous-Time Diffusion Models as Entropy-Regularized Control.
\newblock arXiv:2402.15194.

\bibitem[{Uehara et~al.(2024{\natexlab{b}})Uehara, Zhao, Black, Hajiramezanali, Scalia, Diamant, Tseng, Levine, and Biancalani}]{sergey2}
Uehara, M.; Zhao, Y.; Black, K.; Hajiramezanali, E.; Scalia, G.; Diamant, N.~L.; Tseng, A.~M.; Levine, S.; and Biancalani, T. 2024{\natexlab{b}}.
\newblock Feedback Efficient Online Fine-Tuning of Diffusion Models.
\newblock arXiv:2402.16359.

\bibitem[{Wu et~al.(2023)Wu, Hao, Sun, Chen, Zhu, Zhao, and Li}]{hps}
Wu, X.; Hao, Y.; Sun, K.; Chen, Y.; Zhu, F.; Zhao, R.; and Li, H. 2023.
\newblock Human preference score v2: A solid benchmark for evaluating human preferences of text-to-image synthesis.
\newblock \emph{arXiv preprint arXiv:2306.09341}.

\bibitem[{Xu et~al.(2024)Xu, Liu, Wu, Tong, Li, Ding, Tang, and Dong}]{imagereward}
Xu, J.; Liu, X.; Wu, Y.; Tong, Y.; Li, Q.; Ding, M.; Tang, J.; and Dong, Y. 2024.
\newblock Imagereward: Learning and evaluating human preferences for text-to-image generation.
\newblock \emph{Advances in Neural Information Processing Systems}, 36.

\bibitem[{Yu et~al.(2022)Yu, Xu, Koh, Luong, Baid, Wang, Vasudevan, Ku, Yang, Ayan et~al.}]{parti}
Yu, J.; Xu, Y.; Koh, J.~Y.; Luong, T.; Baid, G.; Wang, Z.; Vasudevan, V.; Ku, A.; Yang, Y.; Ayan, B.~K.; et~al. 2022.
\newblock Scaling autoregressive models for content-rich text-to-image generation.
\newblock \emph{arXiv preprint arXiv:2206.10789}, 2(3): 5.

\end{thebibliography}

\appendix
\newpage
\twocolumn[
\begin{@twocolumnfalse}
\begin{center}
\textbf{\LARGE Supplementary Material \\}
\vspace{3em}
\end{center}
\end{@twocolumnfalse}
]

\section{Dataset details}
In this section, we present the text prompts utilized in the experiments from the main paper. Following baselines~\cite{ddpo, alignprop}, we used prompts in the format of “[animal name]” for experiments with the Aesthetic reward function~\cite{aesthetic} and prompts in the format of “[animal name] [activity]” for experiments with the PickScore reward function~\cite{pickscore}.
We provide the full list of the animal names and activities used to construct the prompts in Table~\ref{table:seen_prompt}.
Additionally, to evaluate generalization ability to unseen prompts, we used additional prompts that were not employed during training. 
Specifically, following the baselines~\cite{ddpo,alignprop}, we used different types of “[animal name]” and “[activity]” as unseen prompts.
We provide the full list of unseen prompts in Table\ref{table:unseen_prompt}.
Lastly, to evaluate the effectiveness of our method on more complicated prompts, we conducted experiments using the DrawBench~\cite{imagen} prompt set. 
For DrawBench, we used 58 challenging prompts categorized under “Conflicting”, “Gary Marcus et al.”, and “Reddit”.

\section{Implementation details}
We conduct experiments by integrating our approach with two reward fine-tuning methods: DDPO~\cite{ddpo} and AlignProp~\cite{alignprop}. Each experiment is based on the implementation of the respective baseline.
In the DDPO experiments, we set the number of denoising steps (i.e. inference steps) to 50, the CFG scale to 5.0, and sample 64 images in each iteration to apply the policy-gradient method. For the policy-gradient update, we use importance sampling with a clip range of 0.0001. In the AlignProp experiments, we set the number of inference steps to 50, the CFG scale to 7.5, and sample 64 images in each iteration to directly backpropagate the gradient of the reward model through the denoising process. 
Regarding the training details, the learning rate is set to 3e-4 in the DDPO experiments and 1e-3 in the AlignProp experiments. The model is updated using AdamW optimizer with $\beta_1=0.9$ and $\beta_2 = 0.99$ in both experiments. 
We use 4 NVIDIA RTX A6000 GPUs for all our experiments.

\section{Additional generated examples}
\begin{figure*}[t]
  \centering
  \includegraphics[width=0.8\linewidth]{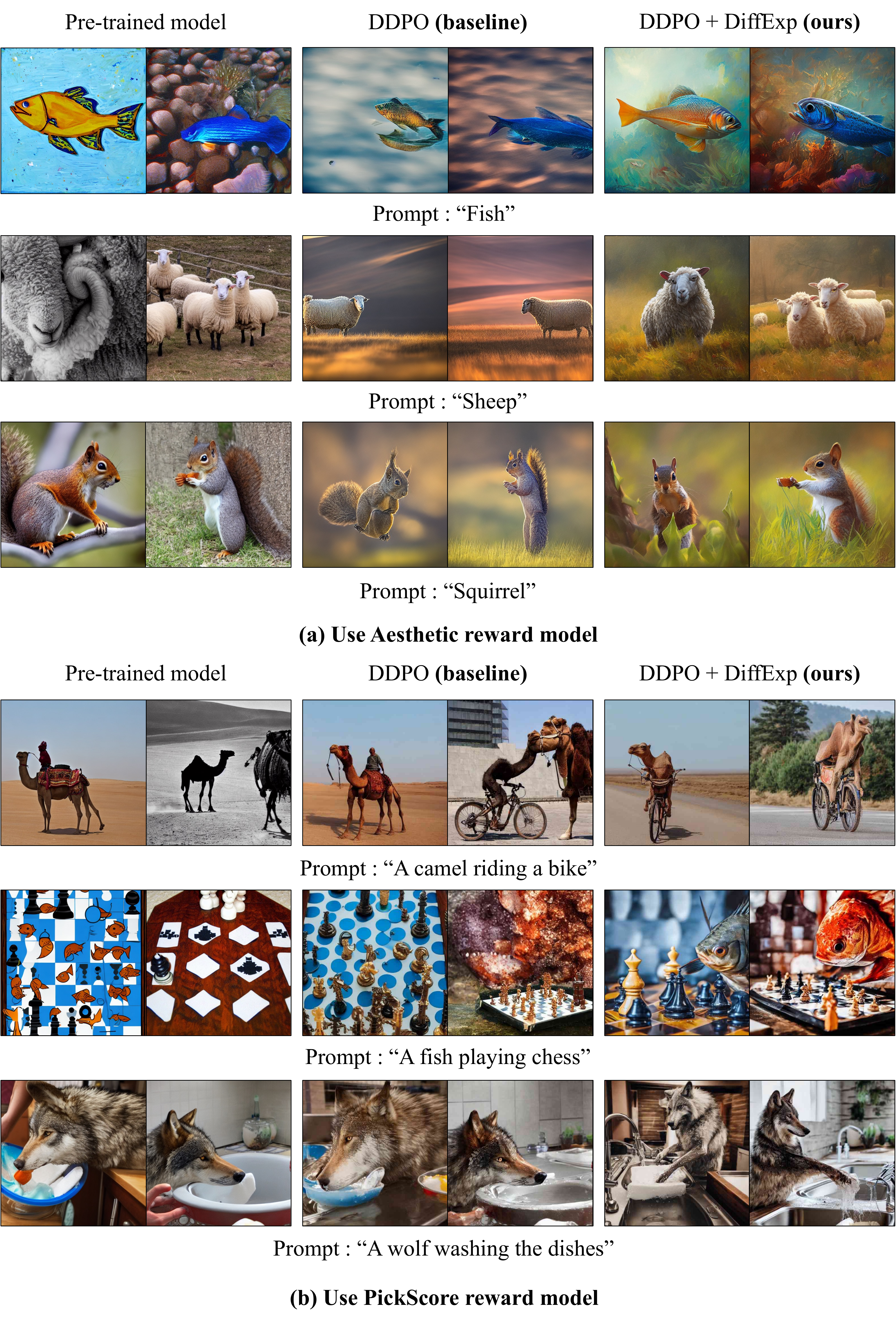}
\caption{Qualitative comparison of images generated from \textit{seen} prompts, using Stable Diffusion v1.5 fine-tuned with DDPO (with or without \metabbr), and without any reward fine-tuning. Images in the same column were generated using the same random seed.}
  \label{fig:ddpo_seen_sup}
\end{figure*}

\begin{figure*}[t]
  \centering
  \includegraphics[width=0.8\linewidth]{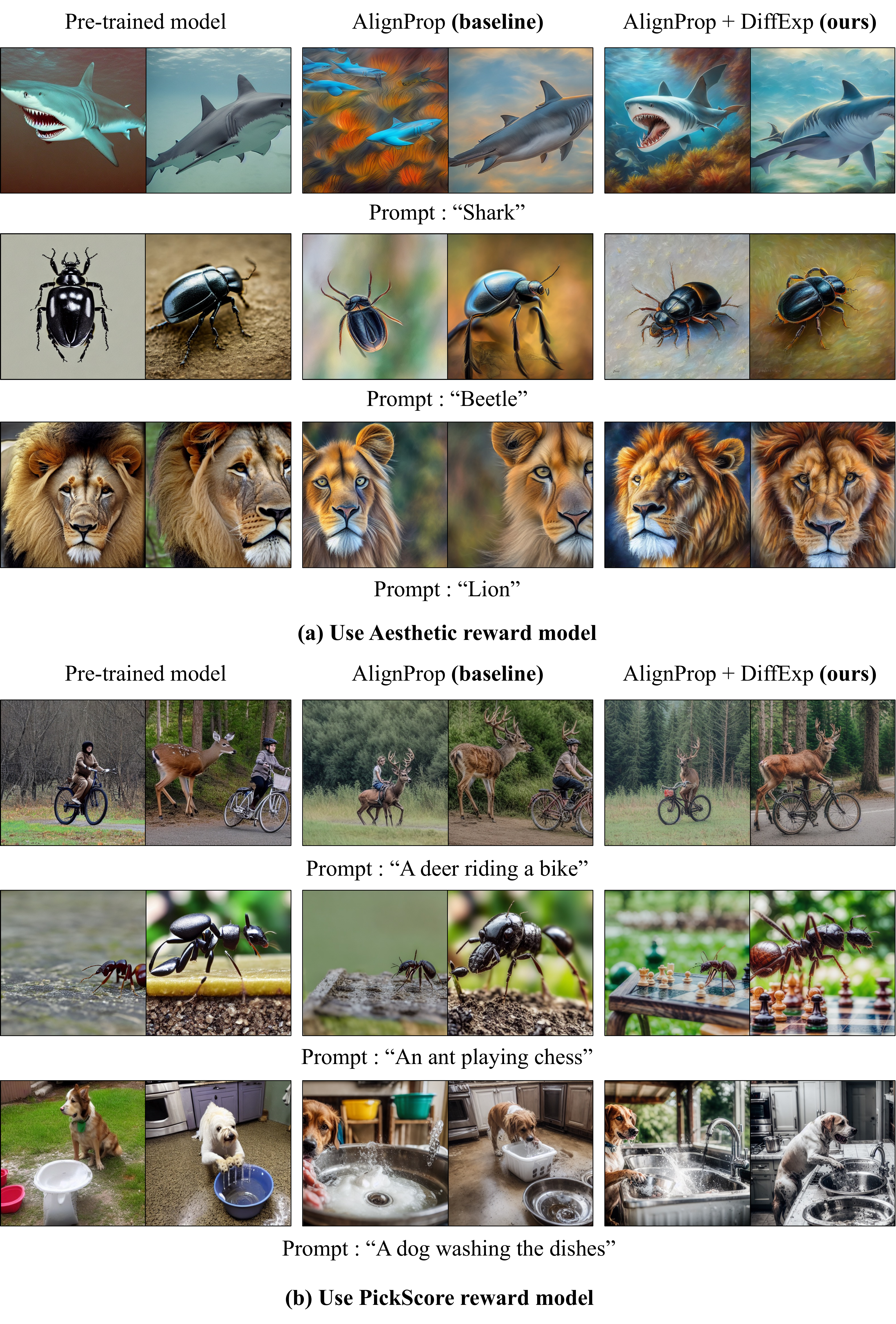}
\caption{Qualitative comparison of images generated from \textit{seen} prompts, using Stable Diffusion v1.5 fine-tuned with AlignProp (with or without \metabbr), and without any reward fine-tuning. Images in the same column were generated using the same random seed.}
  \label{fig:alignprop_seen_sup}
\end{figure*}

\begin{figure*}[t]
  \centering
  \includegraphics[width=0.8\linewidth]{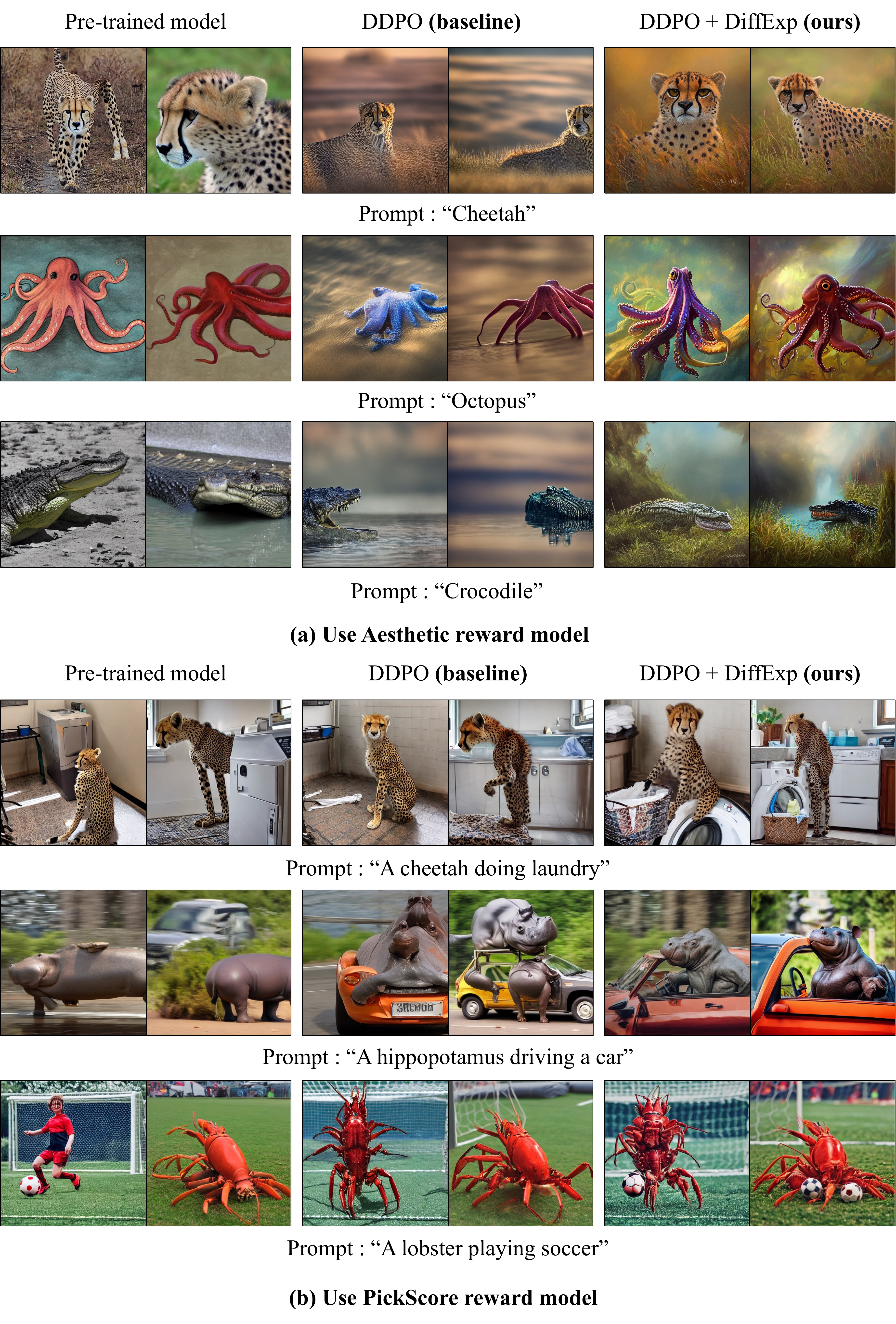}
\caption{Qualitative comparison of images generated from \textit{unseen} prompts, using Stable Diffusion v1.5 fine-tuned with DDPO (with or without \metabbr), and without any reward fine-tuning. Images in the same column were generated using the same random seed.}
  \label{fig:ddpo_unseen_sup}
\end{figure*}

\begin{figure*}[t]
  \centering
  \includegraphics[width=0.8\linewidth]{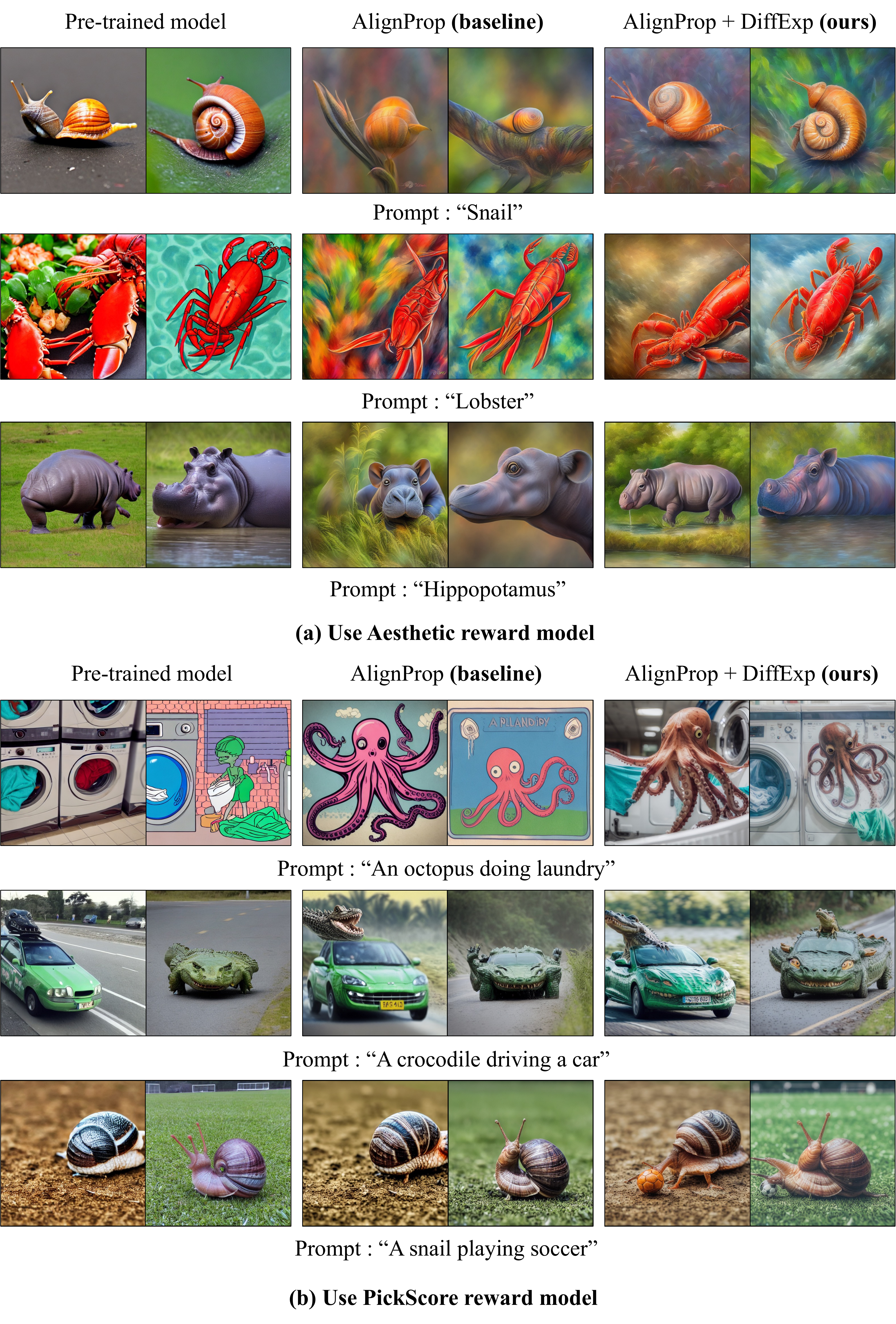}
\caption{Qualitative comparison of images generated from \textit{unseen} prompts, using Stable Diffusion v1.5 fine-tuned with AlignProp (with or without \metabbr), and without any reward fine-tuning. Images in the same column were generated using the same random seed.}
  \label{fig:alignprop_unseen_sup}
\end{figure*}

\begin{figure*}[t]
  \centering
  \includegraphics[width=0.8\linewidth]{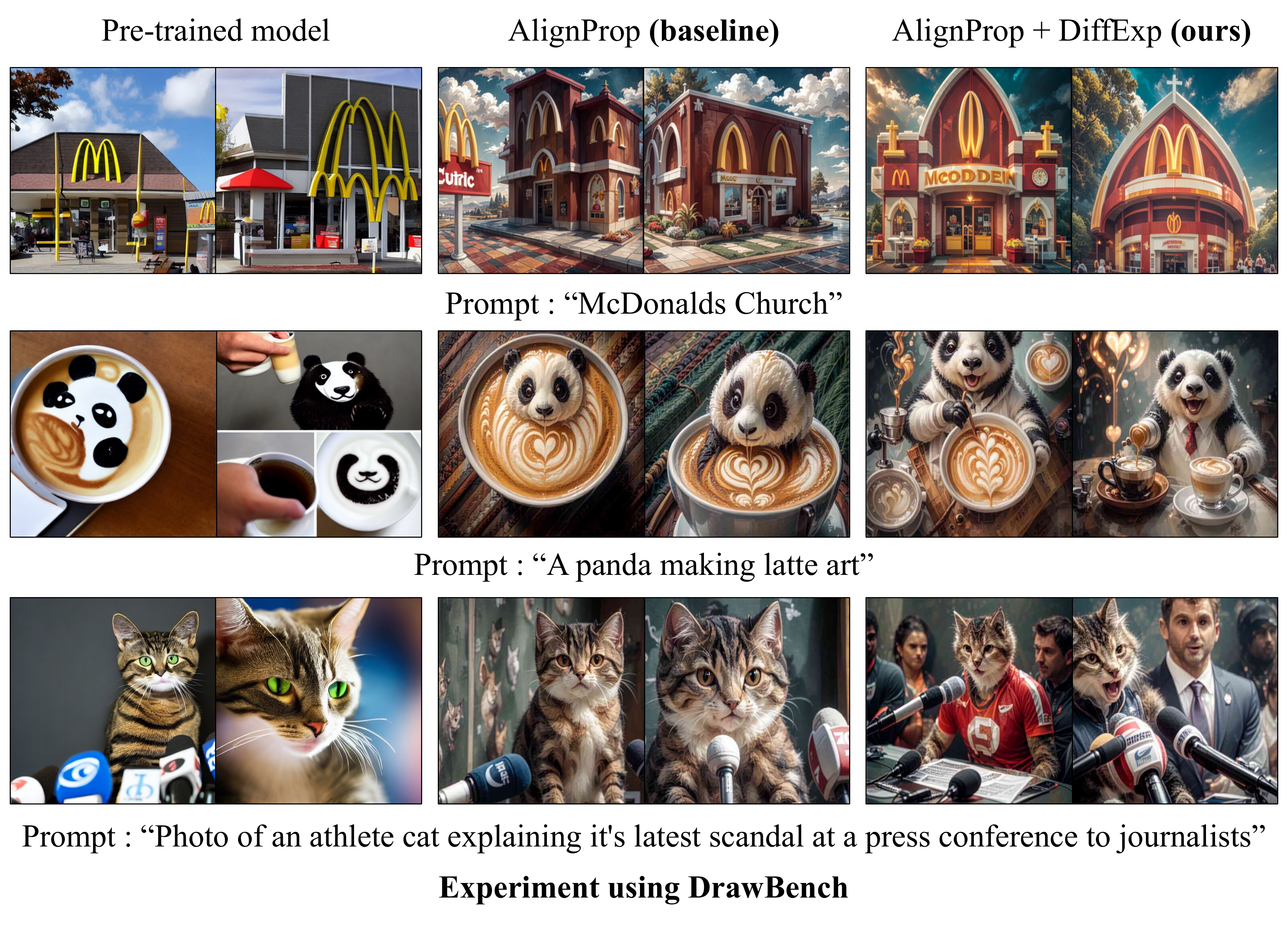}
  \vspace{-1em}
    \caption{Qualitative comparison of images generated from DrawBench prompts (seen), using Stable Diffusion v1.5 fine-tuned with AlignProp (with or without \metabbr) and without any reward fine-tuning. Images in the same column were generated using the same random seed.}
    \vspace{-1.1em}
  \label{fig:drawbench_seen_sup}
\end{figure*}

\begin{figure*}[ht]
  \centering
  \includegraphics[width=0.8\linewidth]{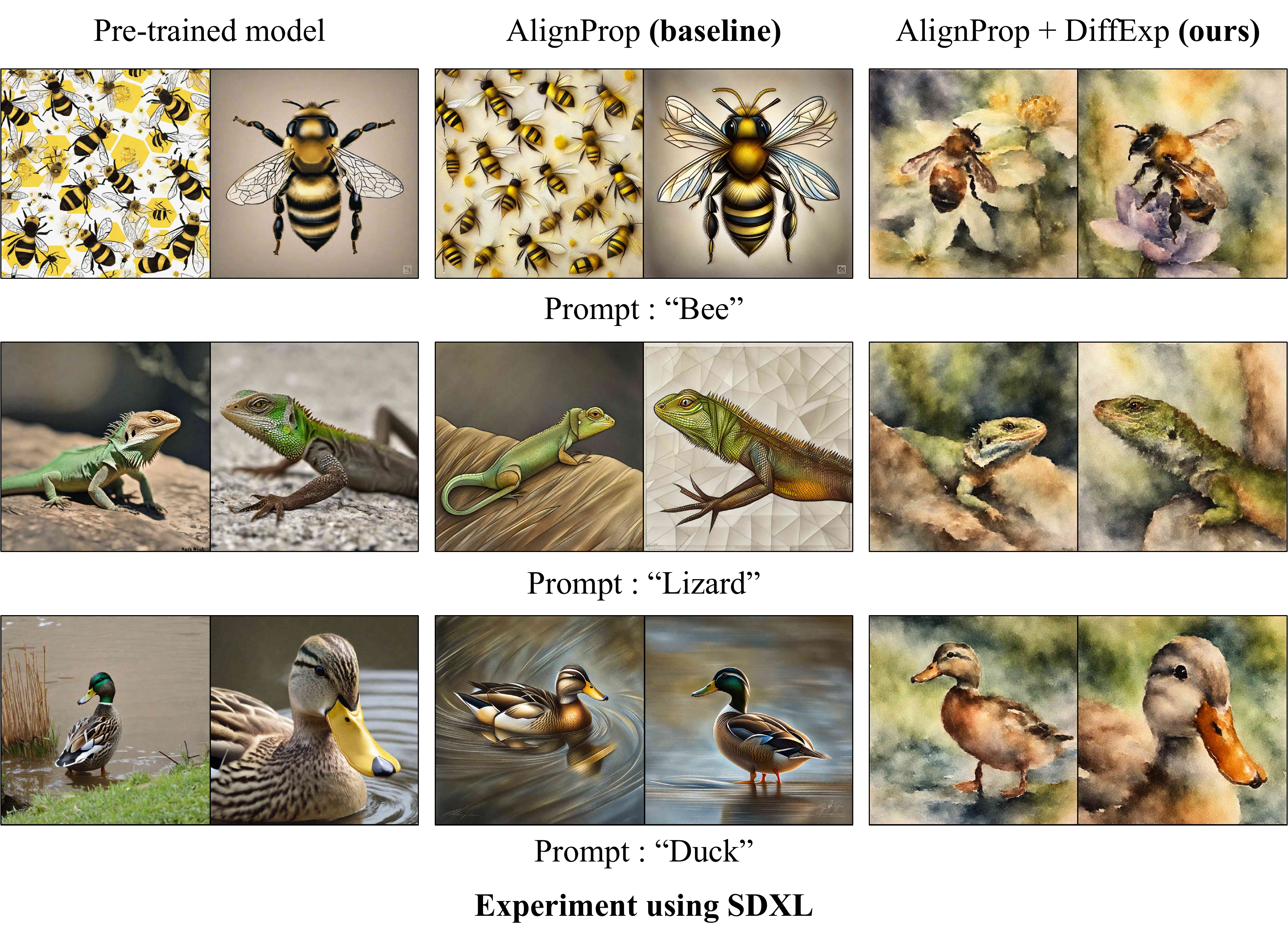}
  \vspace{-1em}
\caption{Qualitative comparison of images generated using SDXL fine-tuned with AlignProp (with or without \metabbr) and without any reward fine-tuning. Note that SDXL is fine-tuned to maximize Aesthetic reward function. Images in the same column were generated using the same random seed.}
  \label{fig:sdxl_seen_sup}
\end{figure*}

We provide additional generated samples comparing our  method to the baselines : DDPO~\cite{ddpo} and AlignProp~\cite{alignprop}. The generated examples using \textit{seen} prompts are presented in Figure~\ref{fig:ddpo_seen_sup} (comparison with DDPO) and Figure~\ref{fig:alignprop_seen_sup} (comparison with AlignProp). We further provide additional samples generated with \textit{unseen} prompts in Figure~\ref{fig:ddpo_unseen_sup} (comparison with DDPO) and Figure~\ref{fig:alignprop_unseen_sup} (comparison with AlignProp).
Finally, in addition to the main experiment, Figure~\ref{fig:drawbench_seen_sup} shows additional images for the experiment with DrawBench~\cite{imagen}, and Figure~\ref{fig:sdxl_seen_sup} presents generated samples from the experiment using SDXL~\cite{sdxl}.

\section{Additional results using unseen prompts}
\subsection{Experiment with DrawBench}
\begin{figure*}[t]
  \centering
  \includegraphics[width=0.8\linewidth]{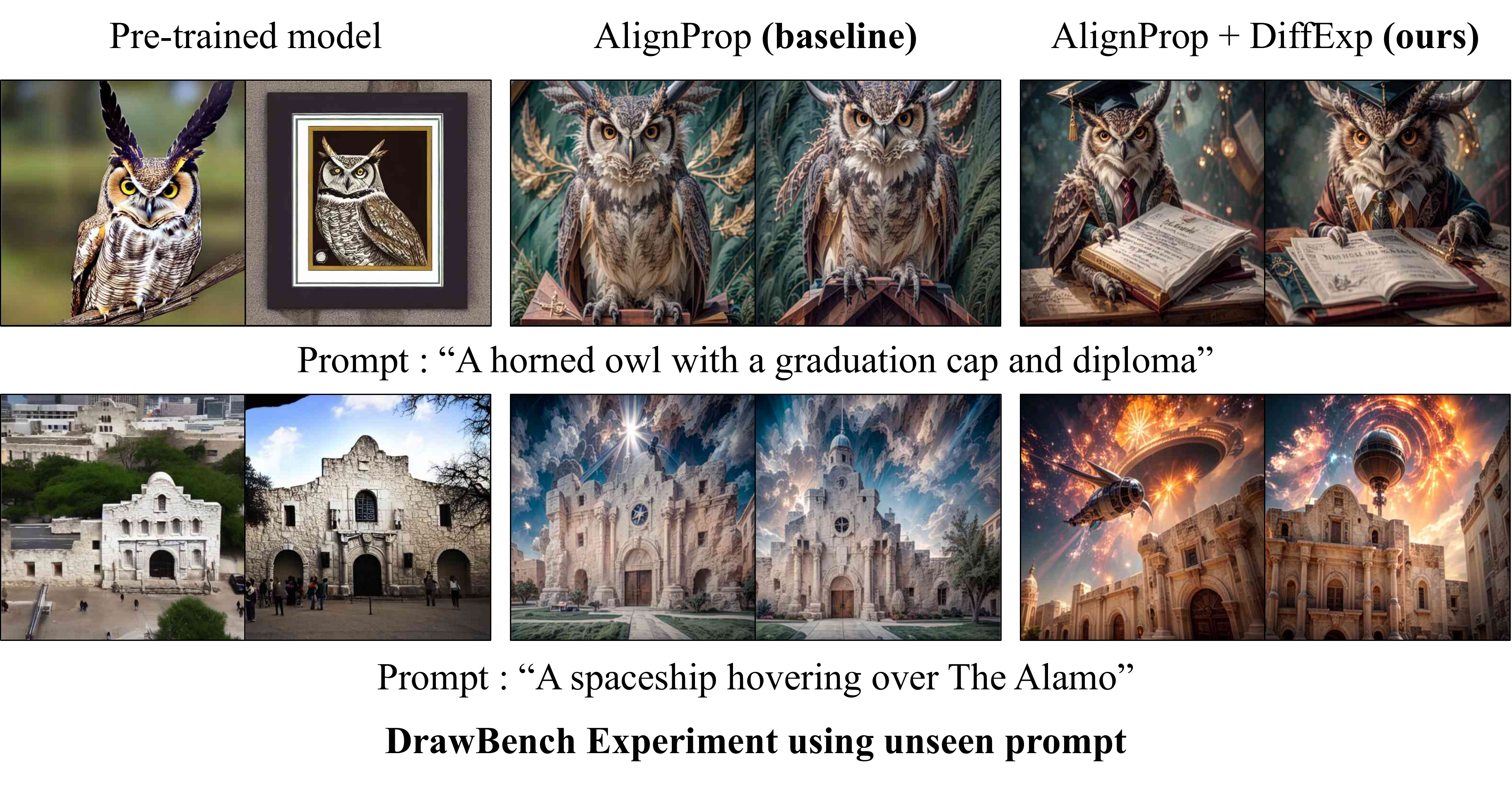}
  \vspace{-1em}
    \caption{Qualitative comparison of images generated from Parti prompts (unseen), using Stable Diffusion v1.5 fine-tuned with AlignProp (with or without \metabbr) and without any reward fine-tuning. Note that the model is fine-tuned using DrawBench, not Parti prompts. Images in the same column were generated using the same random seed.}
  \label{fig:drawbench_unseen_sup}
\end{figure*}

In the main paper, we use the DrawBench~\cite{imagen} prompt set to demonstrate the effectiveness of our method on complex prompts. Here, we further evaluate its generalization ability using 150 challenging prompts from Parti~\cite{parti} as unseen evaluation prompts. The reward curve results for the unseen prompt experiments are presented in Figure~\ref{fig:unseen_reward_curve_sup}(a). 
As shown in Figure~\ref{fig:unseen_reward_curve_sup}(a), our method achieves improved sample efficiency and higher scores compared to the baseline, demonstrating superior generalization ability even on complex prompt sets. We provide the generated images in Figure~\ref{fig:drawbench_unseen_sup}.

\subsection{Experiment with SDXL}
\begin{figure*}[t]
  \centering
  \includegraphics[width=0.75\linewidth]{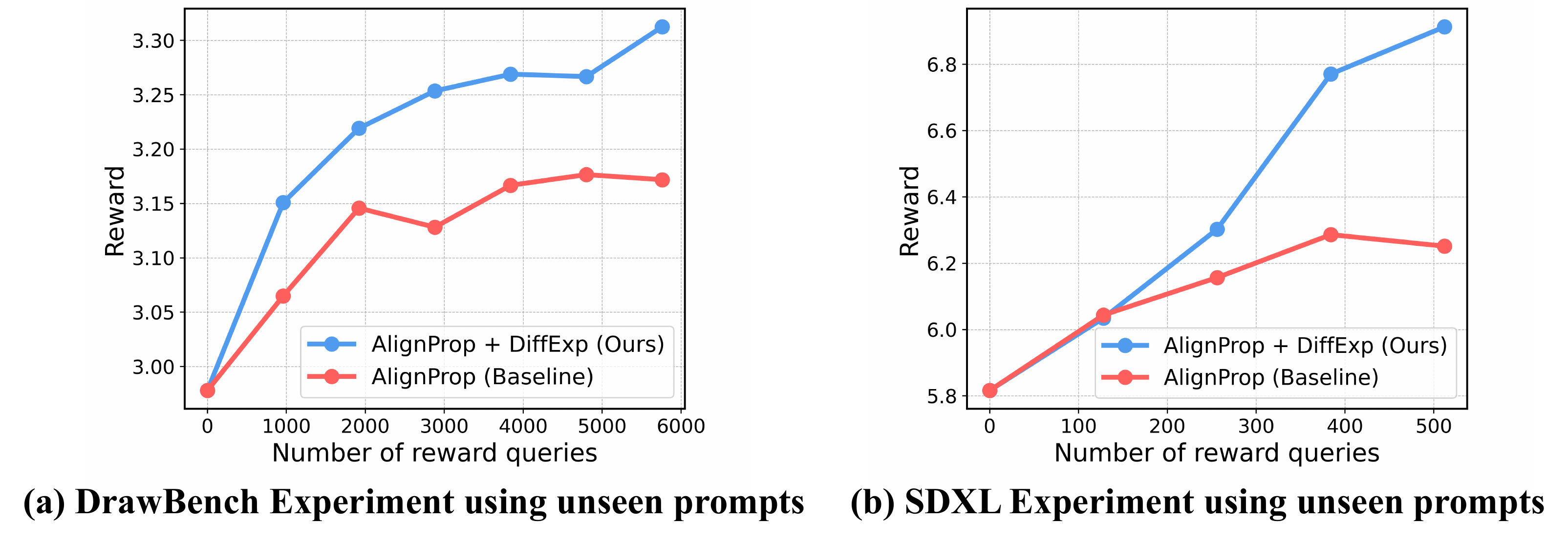}
  \vspace{-1em}
  \caption{Reward curve for unseen prompts. At each checkpoint, we sample 10 images per unseen prompt and calculate the mean of their reward scores to plot the curve. 
  }
  \label{fig:unseen_reward_curve_sup}
  \vspace{-0.5em}
\end{figure*}

\begin{figure*}[t]
  \centering
  \includegraphics[width=0.8\linewidth]{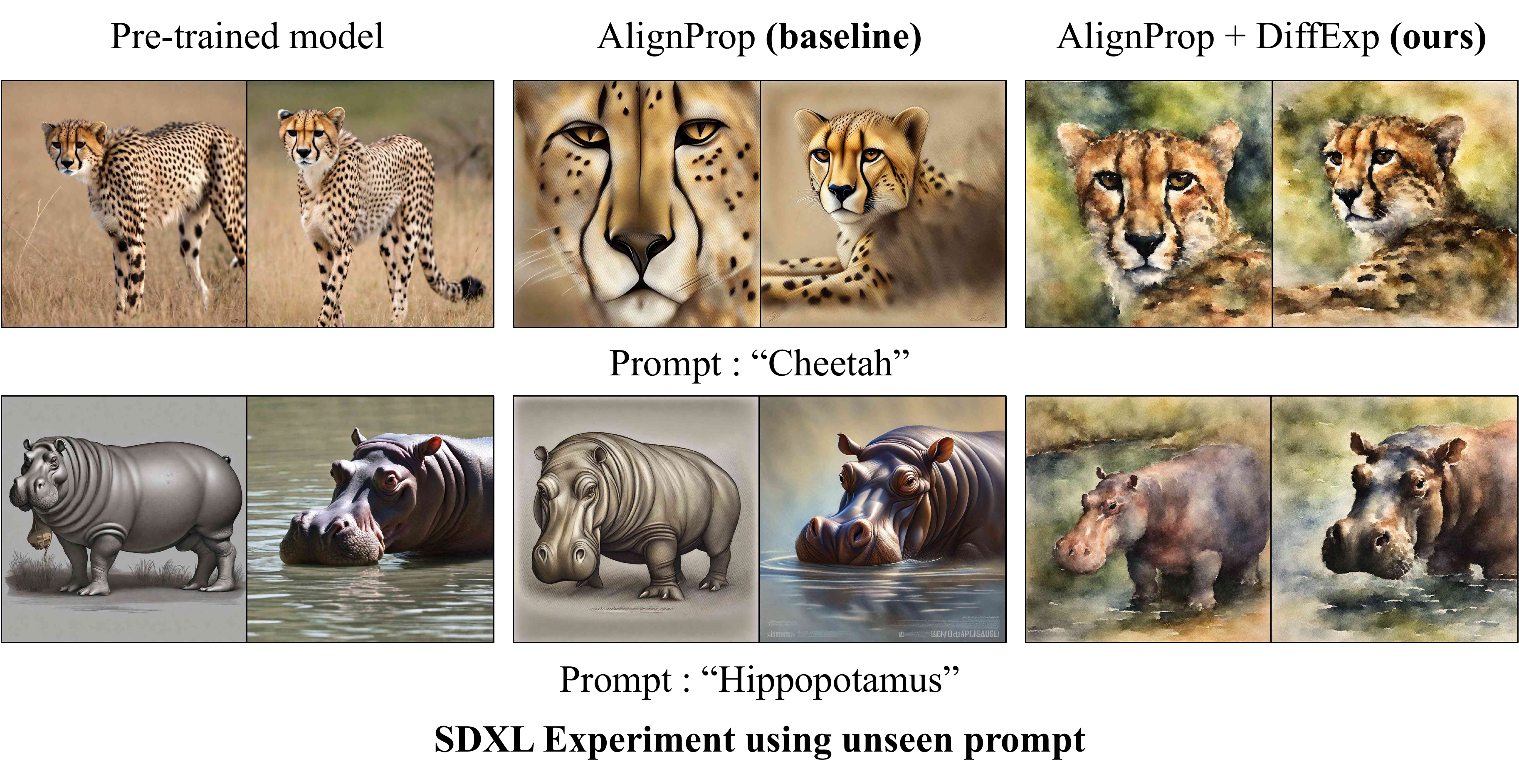}
  \vspace{-1em}
    \caption{Qualitative comparison of images generated using SDXL fine-tuned with AlignProp (with or without \metabbr) and without any reward fine-tuning. Note that SDXL is fine-tuned to maximize Aesthetic reward function. Images in the same column were generated using the same random seed.}
  \label{fig:sdxl_unseen_sup}
\end{figure*}

In the main paper, we demonstrate that our method is also effective when applied to the more advanced diffusion model, Stable Diffusion XL (SDXL)~\cite{sdxl}. 
In this section, we further verify the generalization capability of the SDXL model fine-tuned with our method by testing it on unseen prompts.
Since we fine-tune SDXL using the Aesthetic reward model, we use unseen animal names to construct unseen prompts, which are included in Table~\ref{table:unseen_prompt}.
As shown in Figure~\ref{fig:unseen_reward_curve_sup}(b), our method significantly improves the sample efficiency compared to the baseline, confirming its ability to generalize to unseen prompts even with SDXL.  We provide the generated samples in Figure~\ref{fig:sdxl_unseen_sup}.

\setlength{\tabcolsep}{17pt}
\renewcommand{\arraystretch}{2.0} 
\begin{table*}[ht]
\begin{center}
\caption{The complete list of animal names and activities used for the training prompts in our experiments with Aesthetic~\cite{aesthetic} and PickScore~\cite{pickscore}.}
\label{table:seen_prompt}
\begin{tabular}{cccccc}
\hline
\multicolumn{6}{c}{Animal Name} \\ \hline
\multicolumn{1}{c|}{\texttt{ant}} & \multicolumn{1}{c|}{\texttt{bat}} & \multicolumn{1}{c|}{\texttt{bear}} & \multicolumn{1}{c|}{\texttt{beetle}} & \multicolumn{1}{c|}{\texttt{bee}} & \texttt{bird} \\
\multicolumn{1}{c|}{\texttt{butterfly}} & \multicolumn{1}{c|}{\texttt{camel}} & \multicolumn{1}{c|}{\texttt{cat}} & \multicolumn{1}{c|}{\texttt{chicken}} & \multicolumn{1}{c|}{\texttt{cow}} & \texttt{deer} \\
\multicolumn{1}{c|}{\texttt{dog}} & \multicolumn{1}{c|}{\texttt{dolphin}} & \multicolumn{1}{c|}{\texttt{duck}} & \multicolumn{1}{c|}{\texttt{fish}} & \multicolumn{1}{c|}{\texttt{fly}} & \texttt{fox} \\
\multicolumn{1}{c|}{\texttt{frog}} & \multicolumn{1}{c|}{\texttt{goat}} & \multicolumn{1}{c|}{\texttt{goose}} & \multicolumn{1}{c|}{\texttt{gorilla}} & \multicolumn{1}{c|}{\texttt{hedgehog}} & \texttt{horse} \\
\multicolumn{1}{c|}{\texttt{kangaroo}} & \multicolumn{1}{c|}{\texttt{lion}} & \multicolumn{1}{c|}{\texttt{lizard}} & \multicolumn{1}{c|}{\texttt{llama}} & \multicolumn{1}{c|}{\texttt{monkey}} & \texttt{mouse} \\
\multicolumn{1}{c|}{\texttt{pig}} & \multicolumn{1}{c|}{\texttt{rabbit}} & \multicolumn{1}{c|}{\texttt{raccoon}} & \multicolumn{1}{c|}{\texttt{rat}} & \multicolumn{1}{c|}{\texttt{shark}} & \texttt{sheep} \\
\multicolumn{1}{c|}{\texttt{snake}} & \multicolumn{1}{c|}{\texttt{spider}} & \multicolumn{1}{c|}{\texttt{squirrel}} & \multicolumn{1}{c|}{\texttt{tiger}} & \multicolumn{1}{c|}{\texttt{turkey}} & \texttt{turtle} \\
\multicolumn{1}{c|}{\texttt{whale}} & \multicolumn{1}{c|}{\texttt{wolf}} & \multicolumn{1}{c|}{\texttt{zebra}} & \multicolumn{1}{c|}{\texttt{-}} & \multicolumn{1}{c|}{\texttt{-}} & \texttt{-} \\ \hline
\multicolumn{6}{c}{Activity} \\ \hline
\multicolumn{2}{c|}{\texttt{washing the dishes}} & \multicolumn{2}{c|}{\texttt{riding a bike}} & \multicolumn{2}{c}{\texttt{playing chess}}  
\vspace{-14em}
\end{tabular}

\end{center}
\end{table*}

\setlength{\tabcolsep}{17pt}
\renewcommand{\arraystretch}{2.0} 
\begin{table*}[ht]
\begin{center}
\caption{The complete list of animal names and activities used for the unseen prompts in our experiments with Aesthetic~\cite{aesthetic} and PickScore~\cite{pickscore}.}
\label{table:unseen_prompt}
\begin{tabular}{cccccc}
\hline
\multicolumn{6}{c}{Animal Name} \\ \hline
\multicolumn{1}{c|}{\texttt{hippopotamus}} & \multicolumn{1}{c|}{\texttt{snail}} & \multicolumn{1}{c|}{\texttt{crocodile}} & \multicolumn{1}{c|}{\texttt{cheetah}} & \multicolumn{1}{c|}{\texttt{lobster}} & \texttt{octopus} \\ \hline
\multicolumn{6}{c}{Activity} \\ \hline
\multicolumn{2}{c|}{\texttt{doing laundry}} & \multicolumn{2}{c|}{\texttt{driving a car}} & \multicolumn{2}{c}{\texttt{playing soccer}}    
\end{tabular}

\end{center}
\end{table*}

\clearpage
\newpage

\clearpage
\newpage

\end{document}